\documentclass{article}
\usepackage[english]{babel}
\usepackage{blindtext}
\usepackage{tikz}
\usepackage{amsmath}

\usepackage[numbers]{natbib}

\newcount\draft\draft=0 
\usepackage{listings}
\usepackage{enumerate}
\usepackage{svg}

\usepackage{bm}
\usepackage{soul}
\usepackage[normalem]{ulem}

\newcommand{\N}{\mathcal{N}}

\newcommand{\U}{\mathcal{U}}

\ifcsname G\endcsname%
    \renewcommand{\G}{\mathcal{G}}
\else%
    \newcommand{\G}{\mathcal{G}}
\fi%
\ifcsname C\endcsname%
    
\else%
    
\fi%

\newcommand{\colorcomment}[3]{\textcolor{#1}{[{#2} {#3}]}}
\newcommand{\bluecomment}[2]{\colorcomment{blue}{#1}{#2}}
\newcommand{\redcomment}[2]{\colorcomment{red}{#1}{#2}}
\newcommand{\magentacomment}[2]{\colorcomment{magenta}{#1}{#2}}

\ifnum\draft>2
  \newcommand{\mnote}[1]{\bluecomment{*}{#1}}
  \newcommand{\mcomment}[1]{\magentacomment{Comment by}{#1}}
  \newcommand{\fixthis}[1]{\redcomment{TODO}{#1}}
\else
  \newcommand{\mnote}[1]{\ignorespaces}
  \newcommand{\mcomment}[1]{\ignorespaces}
  \newcommand{\fixthis}[1]{\ignorespaces}
\fi

\ifnum\draft>0
  \newcommand{\preliminarytext}[1]{\textcolor{blue}{#1}}
  
\else 
  \newcommand{\preliminarytext}[1]{#1}
  
\fi

\ifnum\draft>1
  \newcommand{\removedtext}[1]{\textcolor{red}{\sout{#1}}}

\else

  \newcommand{\removedtext}[1]{\ignorespaces}
\fi

\usepackage{tabu}
\usepackage{wrapfig}
\usepackage{graphicx}
\usepackage[all]{xy}
\usepackage{subfigure}

\usepackage{color}
\usepackage{fancyvrb}

\newcommand\blfootnote[1]{%
  \begingroup
  \renewcommand\thefootnote{}\footnote{#1}%
  \addtocounter{footnote}{-1}%
  \endgroup
}

\usepackage{arxiv}

\usepackage[utf8]{inputenc} 
\usepackage[T1]{fontenc}    
\usepackage[hidelinks]{hyperref}       
\usepackage{url}            
\usepackage{booktabs}       
\usepackage{amsfonts}       
\usepackage{nicefrac}       
\usepackage{microtype}      
\usepackage{lipsum}		
\usepackage{graphicx}
\usepackage{natbib}
\usepackage{doi}

\title{Training RL Agents for Multi-Objective Network Defense Tasks}

\date{January 31, 2025}	

\author{\normalfont
  Andres Molina-Markham\textsuperscript{1},
  Luis Robaina\textsuperscript{1},
  Sean Steinle\textsuperscript{1},
  Akash Trivedi\textsuperscript{1},\\
  Derek Tsui\textsuperscript{1},
  Nicholas Potteiger\textsuperscript{1},
  Lauren Brandt\textsuperscript{1},
  Ransom Winder\textsuperscript{1}, 
  Ahmad Ridley\textsuperscript{2}\\\\
  \textsuperscript{1}The MITRE Corporation
  \textsuperscript{2}NSA
}

\hypersetup{
pdftitle={A template for the arxiv style},
pdfsubject={q-bio.NC, q-bio.QM},
pdfauthor={Andres Molina-Markham},
pdfkeywords={First keyword, Second keyword, More},
}

\usepackage{cleveref}
\crefname{section}{§}{§§}
\Crefname{section}{§}{§§}

\begin{document}
\maketitle

\begin{abstract}

Open-ended learning (OEL)~--which emphasizes training agents that achieve broad
capability over narrow competency~--is emerging as a paradigm to develop
artificial intelligence (AI) agents to achieve robustness and generalization.
However, despite promising results that demonstrate the benefits of OEL,
applying OEL to develop autonomous agents for real-world cybersecurity
applications remains a challenge.

We propose a training approach, inspired by OEL, to develop autonomous network
defenders. Our results demonstrate that like in other domains, OEL principles
can translate into more robust and generalizable agents for cyberdefense. To
apply OEL to network defense, it is necessary to address several technical
challenges. Most importantly, it is critical to provide a task representation
approach over a broad universe of tasks that maintains a consistent interface
over goals, rewards and action spaces. This way, the learning agent can train
with varying network conditions, attacker behaviors, and defender goals while
being able to build on previously gained knowledge.

With our tools and results, we aim to fundamentally impact research that applies
AI to solve cybersecurity problems. Specifically, as researchers develop
\emph{gyms} and \emph{benchmarks} for cyberdefense, it is paramount that they
consider diverse tasks with consistent representations, such as those we propose in
our work.

\end{abstract}


\ifnum\draft=0
  \blfootnote{Approved for Public Release; Distribution Unlimited. Public Release Case Number 25-0177. This technical data
deliverable was developed using contract funds under Basic Contract No.
W56KGU-18-D-0004. The view, opinions, and/or findings contained in this report
are those of The MITRE Corporation and should not be construed as an official
Government position, policy, or decision, unless designated by other
documentation. \copyright 2025 The MITRE Corporation. ALL RIGHTS RESERVED.}

\else
  \blfootnote{THIS PRELIMINARY DRAFT IS NOT APPROVED FOR PUBLIC RELEASE. This technical data
deliverable was developed using contract funds under Basic Contract No.
W56KGU-18-D-0004. The view, opinions, and/or findings contained in this report
are those of The MITRE Corporation and should not be construed as an official
Government position, policy, or decision, unless designated by other
documentation. \copyright 2024 The MITRE Corporation. ALL RIGHTS RESERVED.}

\fi

\section{Introduction}

\begin{figure*}[t!]
    \centering
    \subfigure{\includegraphics[trim={0cm 0.6cm 0cm 0.25cm}, clip,width=\textwidth, height=0.37\textheight, keepaspectratio]{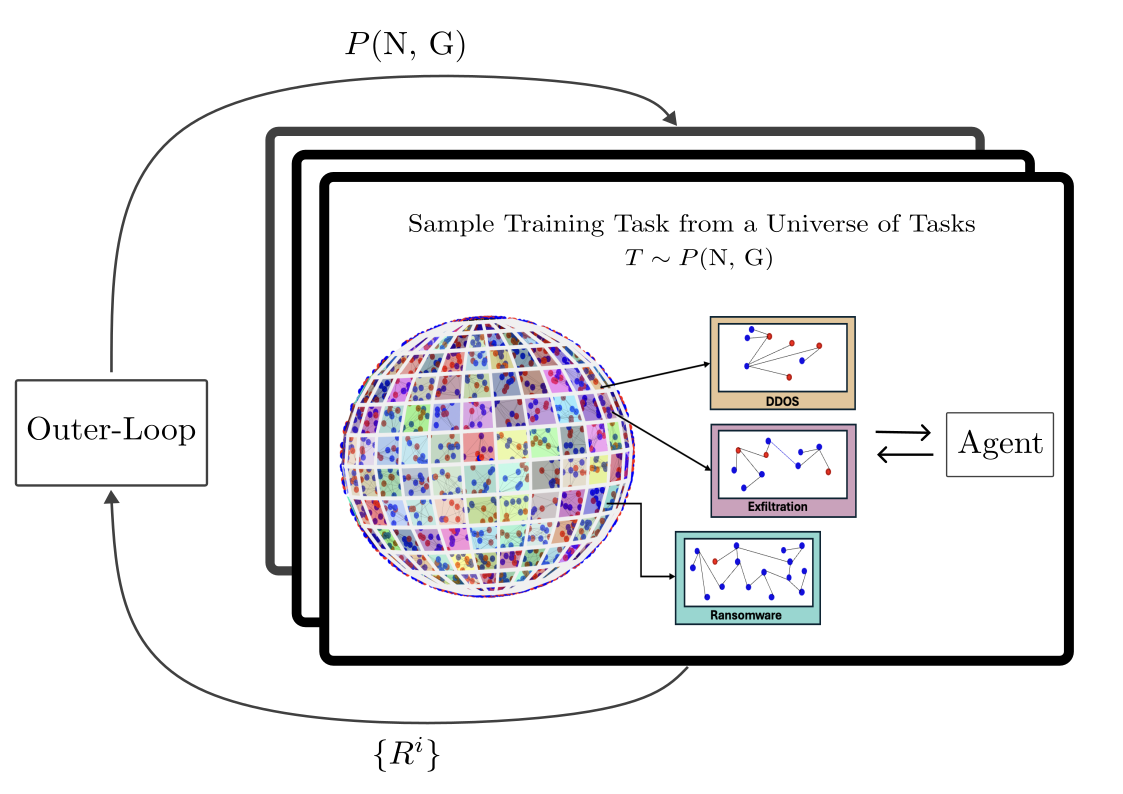}}
\caption{Overview of dynamic task selection: The outer loop adjusts the training-task distribution based on the latest agent policy evaluation, following predefined rules (e.g., a curriculum heuristic that increases task complexity). The agent trains for multiple iterations, with each evaluation round informing the next.}
\label{fig:oel}
\end{figure*}

Reinforcement Learning (RL) has been demonstrated useful to develop agents that
solve complex tasks with high performance. Naturally, this has led researchers
to hypothesize that RL may be used to develop agents to defend the next
generation of computer networks with increasingly complex requirements and
sophisticated adversaries. However, unlike other problems where RL has been
applied successfully, network defense is not characterized by a single task with
a fixed set of rules. Our work explores the applicability of Open-ended Learning
(OEL)~\cite{santucci_intrinsically_2020} to offer a solution to this dichotomy,
enabling autonomous agents to learn a broad set of behaviors that result
in more secure and reliable computer networks, even when information is
imperfect and goals are underspecified.

Open-ended Learning has emerged as a promising learning process for training
agents to achieve multiple goals that require mastering several tasks. Moreover,
OEL expects agents to exhibit useful behaviors when agents face situations that
are unknown at design time~\cite{santucci_intrinsically_2020}. This kind of
learning paradigm is well aligned to the requirements for training agents to
defend networks. Cyberdefenders need to reconfigure networks, balancing multiple
goals and subtasks that must adapt to service demands and evolving threats.
Nonetheless, despite its potential, applying OEL in the context of cyberdefense
is challenging. Crucially, OEL requires adequate representations of goals,
states and actions~\cite{santucci_intrinsically_2020, doncieux_open-ended_2018}
that enable agents to learn complex behavior by becoming competent in a
collection of skills or subtasks. This effectively requires representations and
learning architectures that allow agents to use previously learned skills to
acquire new skills. 
 
Providing a framework that presents a learning agent with a broad set of tasks
is insufficient. If action spaces, state spaces and goal representations are
totally disconnected between two tasks, we can hardly expect that an agent
learns how mastering one task relates to mastering the other. Thus, like
previous work by Stooke et al.~\cite{team_open-ended_2021} in the context of
\emph{physical 3D worlds}, a key component of our approach is to define a
representation for network defense tasks that can be procedurally generated
while preserving the necessary consistency between tasks to enable agents to
learn task relationships. Though previous work has proposed partial answers to
do this effectively in other
domains~\cite{team_open-ended_2021,perez-nieves_modelling_2021,yang_diverse_2021,balduzzi_open-ended_2019,liu_towards_2021},
it is not yet understood how to accomplish this in the context of training
agents to defend computer networks.

Key ideas enabling our approach include the use of action
representations~\cite{chandak_learning_2019} to maintain consistency of action
spaces; and the specification of goals and metrics using the Planning Domain
Definition Language (PDDL). Correspondingly, these two ideas allow agents to
learn consistent action semantics over a task universe; and the representation
of broad sets of goals and metrics that have well-defined relationships.
Importantly, full domain dynamics need
not be specified in PDDL. Defining all aspects of network tasks using PDDL is
intractable for realistic network defense problems. The results of network
defender actions are non-deterministic and preconditions are not fully
observable. 

Adequate representations for network defense tasks must be complemented by an
approach to sample these tasks and present them to the agent during training.
Our work discusses benefits of difficulty-based curriculums (similar to
\emph{automated domain randomization}~\cite{openai_solving_2019}), as well as
the role of task diversity to seek policy robustness and generalization. Similar
to previous work outside network defense, we show that presenting a learning
agent with a collection of subtasks with varying focus and difficulty results in
policies that perform better at solving complex network defense tasks~--when
compared to an approach where the agent interacts with a single environment and
performs a single task (i.e., the end goal).  Arriving at a useful policy with
this approach requires experiencing fewer timesteps, and the resulting policy is
more robust, namely, it performs better under conditions not seen during
training. Our results support the hypothesis that the OEL paradigm may be more
suitable for learning network defense tasks than approaches that only train by
interacting with a single instance of an adversary. 

Our work does not fully answer the general question of what distribution of
tasks will produce the best agent. However, we provide insights about aspects of
tasks and curriculums that are relevant for network defense. Specifically, in
the context of defending computer networks, it is important that the behavior of
defending agents be robust to changes in the network (e.g., size, topology,
congestion); changes in the behavior of normal users (gray agents); the presence
of adversaries with different goals (e.g., ransomware vs stealing data); and
changes (to a degree) in the procedures (behavior) employed by the adversary.

Ultimately, our recommendation is that future developments of \emph{gyms} for network
defense~\cite{baillie_cyborg_2020,team_cyberbattlesim_2021,molina-markham_network_2021,li_cygil_2021,andrew_developing_2022},
consider the problem of broad task representation. Our tools and insights
provide a foundation in that regard. We also advocate for the consideration of
broader sets of network defense tasks when developing benchmarks to evaluate the
robustness of network defenses. The issues of generalization in machine learning
solutions for network security problems is an increasingly relevant
topic~\cite{jacobs_aiml_2022,beltiukov_search_2023}.

\subsection*{Contributions}

The following are the most important contributions of this work:

\begin{itemize}
    \item We present evidence to support the idea that OEL presents a promising
    learning paradigm to develop autonomous network defenders. Importantly, when
    an agent learns to master a broad set of tasks, the knowledge of the agent
    can be generalized to master a task from the same universe of tasks without
    having faced that specific task before.
    \item We present a concrete approach for defining a universe of network
    defense tasks, and demonstrate how this results in training an agent that
    learns a policy that can generalize to a reasonably complex network defense
    task.
    \item We explore the extent to which this kind of ``flexible'' learning may
    be an important step toward learning robust policies for network defense
    whose performance does not fall apart when the adversary deviates from its
    basic behavior. 
\end{itemize}

\section{Related Work}

\preliminarytext{Our work is related to two main research areas: (1) research pursuing techniques to train agents to perform a broad set of tasks; and (2) research focusing on the specific aspects of applying reinforcement learning to network defense. With regard to the first area, our work is related to research in meta-learning~\cite{duan_rl2_2016,eysenbach_diversity_2018,sharma_dynamics-aware_2019,finn_online_2019,kalashnikov_mt-opt_2021}, curriculum learning~\cite{bengio_curriculum_2009,sharma_autonomous_2021} and open ended learning~\cite{santucci_intrinsically_2020, team_open-ended_2021}. With regard to the second area, our contributions are complementary to prior work developing autonomous network defenders~\cite{kanagawa_rogue-gym_2019, baillie_cyborg_2020, standen_cyborg_2021, li_cygil_2021, foley_autonomous_2022} or autonomous red agents~\cite{tran_deep_2021, zhou_autonomous_2021, gangupantulu_using_2022}.}

\preliminarytext{Multiple approaches have been developed, in the context of robotics, with the general goal of training robots to perform complex tasks. Eysenbach et al.~\cite{eysenbach_diversity_2018} studied how to hierarchically compose pretrained skills to learn complex tasks. Finn et al.~\cite{finn_online_2019} introduced the notion of online meta-learning to quickly adapt to new tasks.}

\preliminarytext{Other work has focused on the role of task selection for fast, effective, and robust learning. Our work is more closely related to these works, such as curriculum learning~\cite{bengio_curriculum_2009} and open ended learning~\cite{team_open-ended_2021,perez-nieves_modelling_2021,yang_diverse_2021,balduzzi_open-ended_2019,liu_towards_2021}. We share the general goal of achieving robust and generalizable learning. However, our work has an emphasis on adversarial settings. In adversarial settings, an adversary controls aspects of a task definition (i.e., by controlling aspects of the environment), and therefore we study the extent to which RL policies must be robust to various forms of task variation.}

\preliminarytext{Several research efforts have pursued the general problem of building \emph{gyms}, to train network defenders via reinforcement learning~\cite{kanagawa_rogue-gym_2019, baillie_cyborg_2020, standen_cyborg_2021, li_cygil_2021,molina-markham_network_2021}. Kanagawa et al.~\cite{kanagawa_rogue-gym_2019} touch on the issue of generalization. However, the environments in their Rogue-Gym are based on \emph{roguelike} video games. In contrast, our work has sought to be directly applicable to the problem of defending realistic computer networks. Other works have similarly prioritized relevance to network defense~\cite{baillie_cyborg_2020, standen_cyborg_2021, li_cygil_2021}. CyberBattleSim~\cite{team_cyberbattlesim_2021} was created to be an experimental platform to train automated agents through RL and research and investigate how they operate in enterprise network environments, using a high-level simulated abstraction of computer networks and cybersecurity concepts. The Cyber Operations Research Gym (CybORG)~\cite{baillie_cyborg_2020, standen_cyborg_2021} was an environment developed to support the creation of agents capable of autonomous cyber operations with the expectation of training in simulation and the expressed ambition to address the gap between this and validation for deployment to real environments. CyGIL~\cite{li_cygil_2021} is an architecture providing an environment of network operations for emulated RL training, which also includes the integration of state-of-the-art industry tools to increase agent capability. The Primary-level AI Training Environment (PrimAITE)~\cite{mccarthy_primaite_2023} is a simulator gym for training and assessing AI for cyberdefense against modeled cyberattacks, where training uses RL reward functions balance countering adversaries and maintaining successful operations. Yawning Titan~\cite{andrew_developing_2022} is a simulator that stands as a different approach for simulation toward developing cyber defenses, focusing on models for causal inference, using dynamic causal Bayesian optimization (DCBO) to inform blue actions. FARLAND (the framework for advanced RL for autonomous network defense)~\cite{molina-markham_network_2021}, enables a practical approach to create RL-enabled autonomous network defense agents and the evaluation of their robustness against attacks.} 

\preliminarytext{RL is a typical approach for training defenders with wide possibilities for implementation. In Bates et al.~\cite{bates_reward_2023}, the focus was on reshaping how defenders are rewarded, expanding on the basis of giving penalties for adversarial success to inquire as to the effects of the relative magnitude of penalties, of introducing positive rewards, and of establishing agent curiosity, motivating agent exploration of lesser known states, which determined that the precise techniques used matters greatly to ensure improvements instead of degradation. Research has also examined specific approaches of RL against different cyber use cases. In~\cite{basori_deep_2020}, RL was pitted against other machine learning approaches (CN2, SVM) in an intrusion detection context. In~\cite{hammar_learning_2024}, the use case considered was intrusion response, where the authors employed a simulation system in which to train defenders through their RL method of Threshold Fictitious Self-Play which were in turn evaluated in an emulation system, although this emulation system was designed around intrusion response specifically. }

\preliminarytext{Unlike our work which includes an abstraction that allows transition of simulation to emulation, many approaches remain within a simulation environment. Feng and Xu~\cite{feng_deep_2017} couched cyber state dynamics as a two player zero-sum mathematical game and learned an optimal cyber defense strategy for it using an actor-critic neural network trained via DRL. Applebaum et al.~\cite{applebaum_bridging_2022} performed an analysis of Tabular Q-Learning RL as a baseline algorithmic approach to autonomous cyberdefense in a simulated cyber environment expanding from Microsoft's CyberBattleSim. Zhu et al.~\cite{zhu_effective_2024} provided an approach for network defense agents trained with an algorithm combining DDQN and dueling DQN in the CybORG framework, but with the future intent of expanding into experiments inclusive of emulation. In ~\cite{nyberg_cyber_2022, nyberg_training_2023} RL was used to learn defenses for intrusion response simulations in an environment using attack graphs specified in the Meta Attack Language, where the future ambition was to bridge the gap between simulation and real environments. }

\preliminarytext{Our approach is also related to previous work that suggested defining universes of tasks to train cyberdefenders~\cite{molina-markham_network_2021}. However, that work did not specifically study how dynamic task selection impacts learning results, or how to concretely implement network defense representations that provide consistency during training. Our paper proposes concrete implementation approaches and provides empirical evidence to justify our hypotheses.}

\preliminarytext{Reinforcement learning has also been proposed to drive the behavior of red agents~\cite{team_cyberbattlesim_2021,tran_deep_2021, zhou_autonomous_2021, gangupantulu_using_2022}. Janisch et al.~\cite{janisch_nasimemu_2023} introduced a red-focused gym, Network Attack Simulator \& Emulator (NASimEmu), for DRL training offensive penetration agents that can withstand and adapt to novel scenarios, allowing transfer to new settings with different network topology, size, and configuration and supporting seamless deployment from simulation training to emulation. Like all this work, our work is directly concerned with the practical aspects of applying RL to network defense. However, their rationale and approach are different. Our approach relies on sampling tasks from a universe to achieve generalizability and to facilitate progress during training. In contrast, the goal of red applications is not to find single agents that master multiple tasks. In their approach, it is sufficient to find any number of red agents that induce at least one security violation. }

\preliminarytext{Moreover, research into attacks can extend to examining and mitigating attacks on the controls in cyberphysical systems managed by RL. In Havens et al.~\cite{havens_online_nodate}, to achieve online mitigation of the bias introduced by an attack on a DRL system, the researchers took an approach of optimizing a master policy that can detect attacks and choose between subpolicies appropriate to either the nominal state or the state an adversary perturbed.} 
\section{Approach}

In Section \ref{evaluation}, we discuss the performance of an agent dealing with
a relatively complex network defense task. This agent was trained to master
simpler tasks to both ensure that the agent continues to learn and that we end
with a robust agent that can effectively perform network tasks that are similar
to---but not exactly the same as---a task that the agent has performed before.
This section describes our approach to define network defense tasks and
universes of them. In addition, we discuss our strategy to present new tasks to
a learning agent to achieve both \emph{progress} and \emph{generalization}.

\subsection{Network Defense Tasks}

A network defense task in our framework must adequately capture the security and
quality of service goals of a concrete enterprise network. In other words, a
concrete task encodes various quality of service (QoS) requirements of the
network, as well as the dynamics of the enterprise network, which in turn,
depend on the behavior of the users of the network, the computer applications in
the network, and how the network is configured to process and deliver packets. 

\subsubsection{Task Representation}
\label{subsection:task_representation}

Concretely, our framework defines a network defense task as a pair $(N, G)$, where $N$
defines the network dynamics and $G$ defines the security and liveness goals of
the network defender. The network dynamics $N$ consist minimally of $(I,
\{\pi_{g_i}\}, \{\pi_{r_j}\}, f)$, an initial network configuration; a set of
policies that drive the behavior of gray agents; a set of policies that drive
the behavior of red agents; and, a function that transforms an action
representation, \preliminarytext{which is typically a high-level, abstract description of a defender action},
into an action that can be executed in the simulation. 

$I$ includes a description of the set of initial hosts in the network, how they are connected, what
software is installed in each host, and what packet forwarding rules initially
apply. $I$ also determines where red remote agent tools (RATs) are installed.
Policies in $\{\pi_{g_i}\}$ and $\{\pi_{r_j}\}$ can be implemented via traditional programs,  
stochastic policies, or planners. 

The dynamics of a RL task also depend on the sets of states and actions, the observation model, and
the state transitions. Crucially, $f$ is a key component of the task definition.
As we elaborate in Section \ref{states_and_actions}, to be able to
implement OEL ideas we need stable observation and action spaces. For this, we
rely on action representations~\cite{chandak_learning_2019} that are mapped (by
$f$) to complex actions that can be executed in the environment.

$G$ is specified as a set of conditions that are checked during and at the end
of an episode, and a real valued function (reward function) of these
conditions. For example, when using sparse rewards, $G$ is specified as a
function of the number of times conditions (related to security or quality of
service) were observed during an episode. These conditions may include, among
others, network hosts reporting inability to transfer a file from one host to
another or hosts with a red RAT being captured in a honey network. 
\preliminarytext{Section \ref{subsec:goals} describes goals in our framework in more detail.}

The aspects $(N, G)$ of a network defense task are all aspects that are likely
to change in a real network. For that reason, it is necessary for an agent to
learn robust policies that perform well for most tasks in a universe $\N \times
\G$, for a suitable set of network dynamics $\N$ and a set of security and
liveness goals $\G$ (cf. ~\cref{subsec:goals}). The behavior of authorized users in a
network and the sets of tactics, techniques, and procedures that dictate the
behavior of an adversary can be very broad. However, when developing an RL
network defender, $\U = \N \times \G$ should not be arbitrary. Instead, we argue
that the definition of $\U$ and the selection of tasks in $\U$, during training
and evaluation, should be important considerations for the development of said
defender.

As expected, the performance of a RL network defender also depends on what it can
observe and the actions it can take. 
\preliminarytext{To ensure a consistent interface for the RL defender, varying the task parameters 
$(N, G)$ across the universe of tasks $\N \times \G$ must be done in a way that
maintains a standardized set of possible actions and observations, as described below.}

\subsubsection{Network Dynamics}
\label{states_and_actions}

Similarly to Stooke et al.~\cite{team_open-ended_2021}, when we train a network
defender, we fix the sets of actions $A$ and observations $O$ for a task
universe $\U$.

Fixing $O$ is generally not difficult as representations of state do not change
as frequently. These would only need to be adjusted, for example, when new
sensors or data sources are added. On the other hand, fixing $A$ presents a
technical challenge because defender actions are more accurately characterized 
as parametric actions, such as isolate host with IP 192.168.1.103; re-image host with hostname \emph{webserver1.example.com}; or
start routing packets from host with IP in range $R_1$ to hosts with IP in range
$R_2$ through the network function with IP 192.168.100.1. \preliminarytext{With this parametric representation, the set of defender actions grows as the network scales. For example, a network with 200 hosts results in an action space of about 3,000 possible actions.}

To address this challenge, we use action representations with
heuristics that depend on the state of the network. For example, the actions
above could be described as non-parametric actions, such as:
\emph{isolate-host-flagged-by-nids}; \emph{reimage-webserver}; or
\emph{start-deep-packet-inspection-in-zone-flagged-by-nids}. Because there is no
unique way to transform these non-parametric actions into parametric actions,
our approach requires specifying such transformation $f$ in the definition of
the dynamics $N$ of the network task.

Note that fixing actions and observation spaces does not mean that the dynamics
are fixed. The initial conditions and the policies of gray and red agents $(I,
\{\pi_{g_i}\}, \{\pi_{r_j}\}$) can still vary during training. In addition,
different network defense tasks in a curriculum can also specify different goals
for the defender agent, as we describe next.

\subsubsection{Goals}
\label{subsec:goals}

One typical way to encode the targeted behavior of RL agents is through the
specification of reward functions. However, it is not feasible to manually
tailor reward functions to encode each behavior in a large set of desired
behaviors. This presents a technical challenge that we address by encoding a
reward as a function of a \emph{goal} and a \emph{metric} that helps
to optimize for desirable characteristics of solutions. Specifically, 
in our framework, each network defense task defines a boolean goal and a numeric 
metric that combine terms using PDDL syntax. These terms describe a
goal that must be satisfied at the end of an episode; and, a metric that is
evaluated at the end of the episode (when using a sparse reward) or every step
(when using a dense reward).

With these goal-metric pairs we defined a sparse reward function as follows,
where $x$ is the result of evaluating the numeric expression corresponding to
the \emph{metric}, and $g$ is the boolean expression corresponding to the
\emph{goal}:
\begin{equation*}
  r(x) = \left\{ \begin{matrix}
    \frac{1}{2e^{x^2}} + 0.5 & \text{if $g$ is satisfied}  \\
    \frac{1}{2e^{x^2}} & \text{if $g$ is not satisfied}  
\end{matrix} \right.
\end{equation*}

\begin{figure}[ht]
  \centering
  \includegraphics[width=0.5\textwidth]{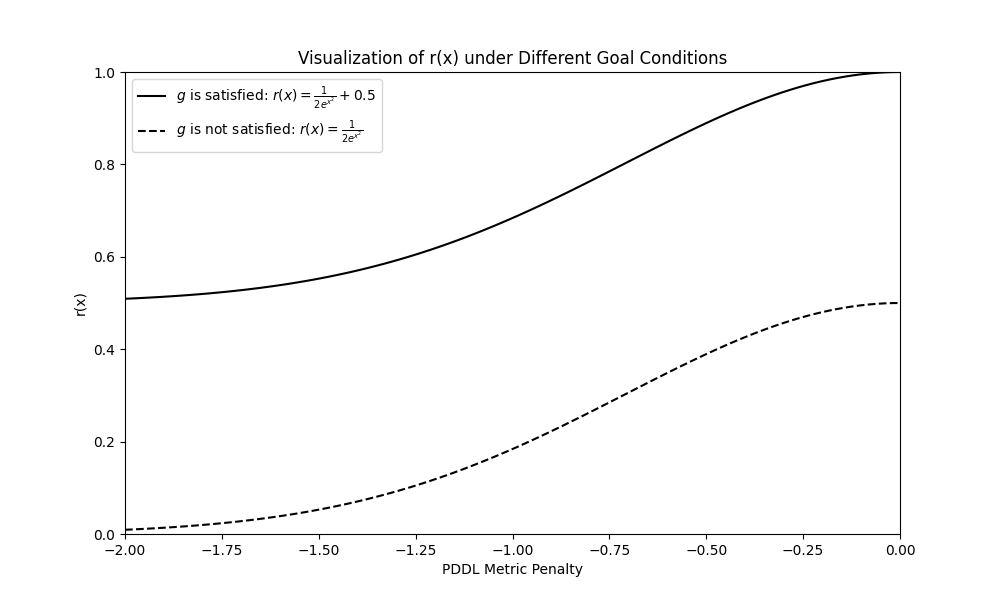}
  \caption{\preliminarytext{Reward curve derived from the goal-metric pair shown in Figure~\ref{fig:goals}.}}
  \label{fig:r_x_plot}
\end{figure}

It is possible to specify other reward functions from a goal-metric pair. \preliminarytext{We selected this approach because it provides intuitive and desirable properties. The reward function is bounded between 0 and 1, where 1 represents the most desirable outcome. When \(x\) is close to zero (indicating less penalties), \(r(x)\) approaches 1 if the goal is satisfied and at most 0.5 if the goal is not satisfied. Conversely, as \(x\) decreases (indicating more penalties), \(r(x)\) converges to 0.5 if the goal is satisfied and to 0 if the goal is not satisfied. This approach offers a more intuitive and interpretable reward signal compared to other methods, where reward values may drop to extreme ranges \cite{cage_challenge}, which can be difficult to interpret. For instance, using the definition of $r(x)$ and the example goal-metric pair shown in Figure \ref{fig:goals}, we derive the reward curve illustrated in Figure~\ref{fig:r_x_plot}.}

In our evaluation, the goal-metric pairs are expressed as ratios. For metrics that yield larger integer values for \(x\), alternative reward functions may be more appropriate to prevent \(r(x)\) from dropping to zero too quickly. 

\subsubsection{The Use of PDDL}
\label{subsec:pddl}

\preliminarytext{Given PDDL's well-established role in defining planning problems and domains for symbolic planning and problem solving \cite{zhi_xuan_2022}, we build on this familiarity for task representation in our approach. PDDL enables us to formally specify goals and metrics, mask actions, and create plans for bootstrapping. By leveraging PDDL, we gain a consistent and flexible approach to defining these core components of our network defense tasks, which in turn enhances training efficiency and supports the development of robust RL agents. Below, we detail the three key areas where PDDL is used in this work:}

\paragraph{\textbf{Goal-Based Rewards:}}
\preliminarytext{As mentioned in Section \ref{subsec:goals}, manually defining reward functions for a wide range of behaviors is impractical. Although it is possible to use programming languages like Julia to implement reward functions by parsing agent actions and observed states, this approach often presents several challenges. For example, as the action and state spaces grow, parsing these actions and states into meaningful numeric rewards becomes increasingly cumbersome, making the system difficult to scale.} 

\preliminarytext{To efficiently define a diverse set of goals and metrics, we instead use PDDL. The key advantage of using PDDL is the ability to crisply specify a broad range of goals and behaviors while minimizing complexity and scaling issues that come with parsing state-action data manually. Additionally, PDDL simplifies the process of adapting reward functions as new tasks or behaviors emerge.}

\preliminarytext{We utilize PDDL to define boolean goals and numeric metrics for each network defense task, forming the core of our reward structure. The goal must be satisfied by the end of an episode, while the metric is evaluated either at the episode's conclusion (for sparse rewards) or at each step (for dense rewards). Figure \ref{fig:goals} illustrates one example of a goal-metric pair. Additional examples and a comprehensive list of the terms used in our evaluation are available in Appendix \ref{appendix:pddl_terms}.}

\begin{figure}[ht]
  \centering
  \noindent
  \begin{tabular}{c}
    \begin{lstlisting}
(:goal 
  (or 
    (and (not red-inactive) 
      (not real-compromise) 
      (declared-victory)) 
    (and (red-inactive) 
      (= nontrivial-blue-actions 0))))
(:metric minimize 
  (+ 
    (/ nontrivial-blue-actions 
      steps-to-survive) 
    (/ bad-qos-events 
      (+ bad-qos-events 
        good-qos-events))))
    \end{lstlisting}
  \end{tabular}
  \caption{Example of a goal and a metric in PDDL: When an active red agent is detected, the agent should use decoys as a countermeasure to deceive the red agent into thinking it has successfully compromised a real asset on the network. This behavior is encoded by the requirement that the red agent must "declare victory" without the presence of a "real compromise." In the absence of an active red agent, the agent should remain passive ("do nothing" action). 
  The metric specifies to minimize the ratio of non-trivial actions over the total number of actions, and
  the ratio of negative QoS events over the total number of QoS events.}
  \label{fig:goals}
  
\end{figure}

\paragraph{\textbf{Action Masking:}}
\preliminarytext{The second application of PDDL is in action masking, which limits the RL agent’s action space to only those actions that are actually available in the current state. Without this mechanism, the agent would have access to the entire action space $A$ (described in section \ref{states_and_actions}), which is large and likely to include many actions that are not actually applicable to the current state. Many other RL network defense projects implement action filtering in a rule-based manner \cite{li_cygil_2022, team_cyberbattlesim_2021, standen_cyborg_2021, baillie_cyborg_2020}, where constraints on the available actions are manually defined based on the current state and environmental conditions. These rule-based methods, though effective in some settings, rely on hardcoded constraints that are specific to the environment, making them less scalable and adaptable to more complex or varied scenarios.} 

\preliminarytext{While specifying the full domain dynamics in PDDL is not necessary, defining action preconditions in PDDL enables the use of its built-in functions to dynamically filter out inapplicable actions. This reduces the action set without needing manually defined rules for every possible constraint. As a result, this approach offers a more scalable and flexible solution for action filtering, well-suited for managing complex and dynamic environments. By focusing on actions that can yield meaningful effects, this approach improves the agent’s decision-making and accelerates the learning process.}
 
\paragraph{\textbf{Blue Agent Planning:}} 
\preliminarytext{The third key use of PDDL, which will be discussed in more detail in future work, is blue-agent planning. In this approach, PDDL is used to combine goal-based rewards and masked actions to generate example plans for the blue agent. These plans are then bootstrapped into the RL algorithm, providing the agent with valuable learning experiences that guide its training. By leveraging these PDDL structures, we can create plans that accelerate the agent’s training and enhance its performance. Future work will explore and elaborate on additional aspects and impacts of this planning.}

\subsection{Task Selection}
\label{subsec:task_selection}

There are several challenging questions related to task selection, and, while we
do not offer complete solutions in this paper, we hope that our framework and
insights guide future work. First, it is important to know what strategy should
guide the selection of tasks during training, to achieve optimal progress.
Similarly, what task selection strategy would lead to greater generalization?
The answers to both most likely depend on the training algorithm that is used.
An answer to the first question would result in the most efficient curriculum to
solve specific tasks. An answer to the second question would yield a strategy to
train more \emph{generally capable} agents. Finally, since our goal is to train
robust network defenders, we also care about a third question: How do we select
\emph{testing tasks}? Here, we outline our current task selection strategies and
then we evaluate them in Section \ref{evaluation}. 

Our approach for task selection is loosely related to Stooke et
al.'s~\cite{team_open-ended_2021} and to Akkaya et al.'s
~\cite{openai_solving_2019}. Stooke et al. varied three properties of
tasks~--applied to \emph{worlds} and \emph{agent goals}~--as agents learned:
\emph{smoothness}, \emph{vastness}, and \emph{diversity}. In contrast, Akkaya et
al. only sought to gradually increase the difficulty of tasks during training.
Our notion of smoothness is similar to Stooke et al.'s. However, our 
research motivation includes more modest notions of \emph{diversity} and
\emph{vastness}. As we present new tasks to learning agents, we keep small
distances between tasks (within the universe of tasks). 

Concretely, these are the attributes of network dynamics that may vary during
training:

\begin{enumerate}
    \item Size of a network controlled by two parameters: the number of hosts in
    a network and the number of subnets
    \item Number of initially compromised hosts (with a red RAT from the
    beginning)
    \item The basic behavior of a red agent consistent with one of the following
    goals: (Exfiltration, Ransomware, DDoS, or DoS)
    \item Deviation from a known basic behavior along two dimensions: the time
    between actions consistent with the known basic behavior; and the amount of
    gray-like behavior driven by the red agent aimed at masking red behavior\footnote{We
    capture these two dimensions via parameters of probability distributions
    that affect the sampling of red actions.}
    \item The number of time steps that the blue agent must survive without
    incident 
    \item A goal-metric pair in a set of 43 pairs that range from simply
    detecting the presence of an attack, using specific actions to mitigate the
    attack, to ultimately determining how to efficiently use any available actions
    to maximize QoS and satisfy more sophisticated security goals
\end{enumerate}

The variation of these parameters, whose list can be naturally extended, helps
to provide a somewhat diverse universe of tasks. However, while the number of
possible instantiations of tasks is large, we cannot currently consider $\N$
vast. We leave for future work extending our architecture to draw many more TTPs
from frameworks, such as ATT\&CK \cite{strom_mitre_2018}, and considering richer
forms of injecting normal behavior.

\begin{figure}[!h]
  \centering
  \includegraphics[width=0.5\linewidth]{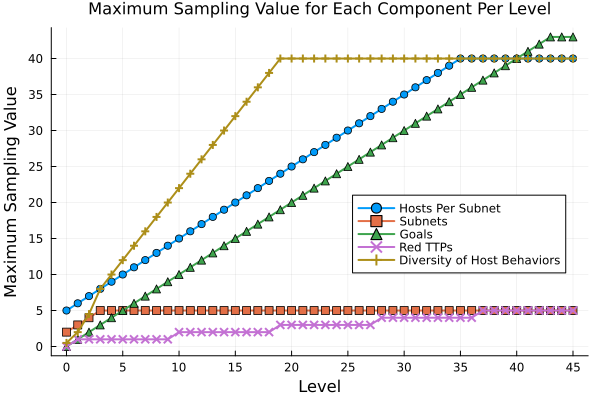}
  \caption{Maximum sampling values at different levels for dynamic task selection of a blue agent with no DoS. Parameters are sampled from a range between the maximum value at each level and the minimum (value at level 0). Diversity of host behavior is calculated as gray traffic divided by two, then multiplied by gray agent diversity, with the division preventing plot flattening at larger y-axis values.}
  \label{fig:bylevel_visual}
\end{figure}

Our task selection strategy, during training, is to simply increase the numbers
(or cardinalities of sets) for each of the parameters that determine $\N$ as
agents' mean scores reach a level and to decrease these numbers as the agents'
mean scores fall below a level. \preliminarytext{Figure \ref{fig:bylevel_visual} provides an example visualization of how the cardinalities for each parameter change as levels increase.} 

To evaluate generalization, we present trained agents with a new task that they
never saw during training: A red agent with a DoS goal and multiple
levels of deviation from its basic behavior. We describe further details about
the testing task in Section \ref{evaluation}. 

\section{Evaluation}
\label{evaluation}

To assess the extent to which OEL may be a suitable learning paradigm to train
autonomous cyberdefenders, we evaluate the following aspects. First, Section
\ref{subsec:scaling} evaluates the benefits of training with a broad universe of
tasks, gradually scaling the complexity of the defense tasks that are presented
to the learning agent. Then, Section \ref{subsec:role_of_difficulty} compares
different strategies for dynamic task selection. Finally, Section
\ref{subsec:generalization} evaluates the degree to which it may be possible to
achieve generalization and robustness via OEL. 

This section also discusses relevant implementation details. Specifically,
Section \ref{subsec:curriculus_with_goals} explains how we created curriculums
with multiple goal-metric pairs. We describe how we estimate the relative
difficulty of each goal-metric as well as how it is possible to use different
approaches for representing goals as observations to guide the learning agent.
Sections \ref{subsec:training} and \ref{subsection:algorithm} also provide
details about the algorithms and hyperparameters that we used in our
experiments\footnote{Source code to replicate our results is forthcoming.}. 

\subsection{Scaling}
\label{subsec:scaling}

With regards to scaling, we evaluate the extent to which using dynamic task
selection during training results in a policy with higher performance and with
shorter training~--when compared to an approach that only trains to master a
task of fixed difficulty.

Specifically, we compare the performance of two policies obtained with two
different task selection strategies, as the agent is learning:
\begin{description}
    \item[Fixed task selection] In this case, the task definition remains
    constant throughout the training period using only a task that is
    representative of the testing task.
    \item[Dynamic task selection] This strategy consists of dynamically
    adjusting the difficulty of the task, as the agent is learning, and until it
    eventually faces the target task.
\end{description}


    
        
            
        
    


\begin{figure}[]
    \centering
    \subfigure{\includegraphics[width=.5\textwidth]{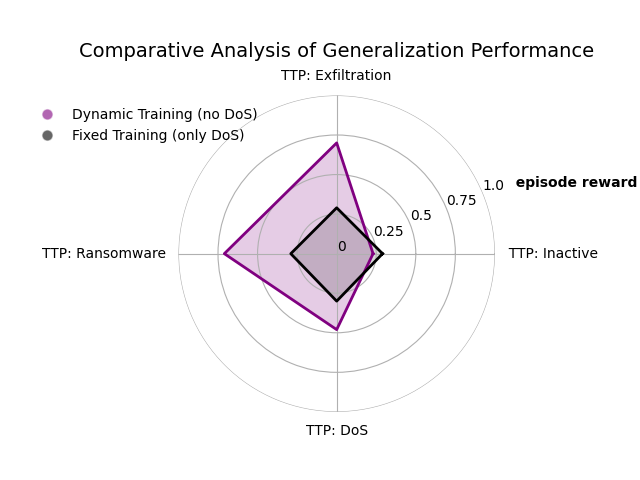}}
    \caption{\preliminarytext{Comparing Defender Generalization Ability: Dynamic vs. Fixed Task Training Approaches}}
    \label{fig:testing_performance}
\end{figure}

\begin{figure*}[t!]
    \centering
\begin{tabular}{ccc}
    \includegraphics[trim={0 0 3.1cm 0},clip,width=.275\textwidth]{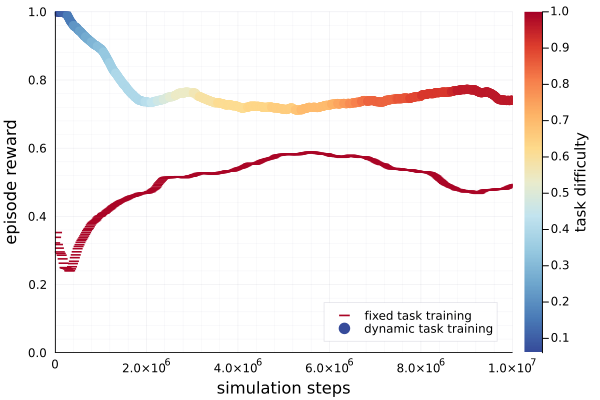} &
    \includegraphics[trim={0 0 3.1cm 0},clip,width=.275\textwidth]{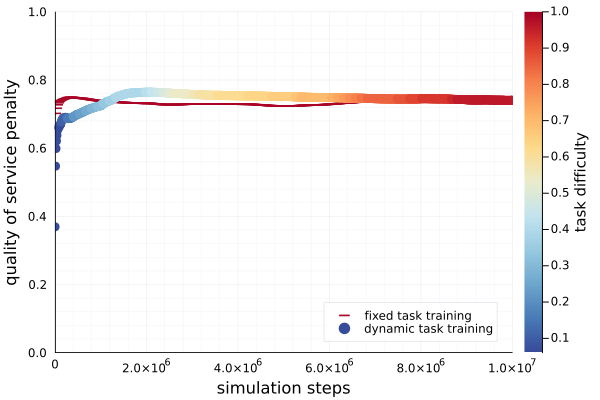} &
    \includegraphics[trim={0 0 0 0},clip,width=.325\textwidth]{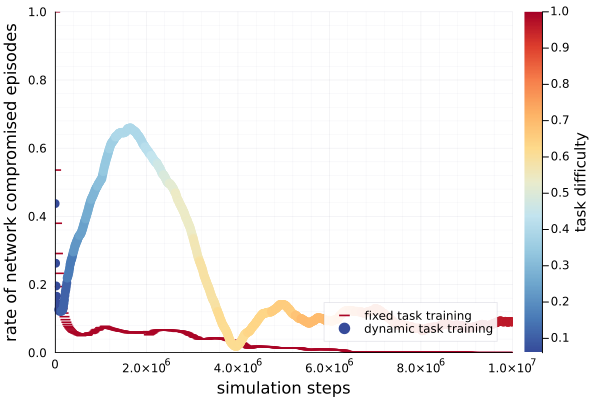} 
\end{tabular}
\caption{Comparison of two policies: one with the fixed task selection strategy from Section~\ref{subsec:scaling}, and another with dynamic task selection (also from Section~\ref{subsec:scaling}). The marker colors indicate task difficulty during training, based on factors like hosts, subnets, and red/gray behavior variability. Figure~\ref{fig:fixed_vs_dynamic_full} shows the comparisons across additional network components.}
\label{fig:fixed_vs_dynamic}
\end{figure*}

Figure \ref{fig:fixed_vs_dynamic} compares the performance of policies with the
two approaches, on multiple dimensions, as the agent trains. Dynamic task
selection outperforms fixed task selection in many regards. Importantly, we can
arrive at a higher performance with less training on a more diverse set of
conditions. 

We also evaluate the performance of the policies with several \emph{testing
tasks} after training has been completed. The testing tasks consist of defending
against one of several attacks in a network with 200 hosts during 100 episodes,
while maintaining a minimum quality of service. The policies of the red and gray
agents are probabilistic, but the corresponding probability distributions
(driving the behavior of the red and gray agents) are fixed. Figure
 \ref{fig:testing_performance} compares the testing performance of policies
after 5M training simulation steps. On tasks at the highest difficulty, policies
trained via dynamic task selection perform better at with regards to mitigating attacks. However, training with fixed task selection provides higher QoS when there is not an active attack.
Policies trained via dynamic task selection outperform those via fixed task
selection in robustness and generalization, as we explain the next sections. 



\label{subsec:role_of_difficulty}

\begin{figure*}[t!]
    \centering
\begin{tabular}{ccc}
    \includegraphics[trim={0 0 3.1cm 0},clip,width=.275\textwidth]{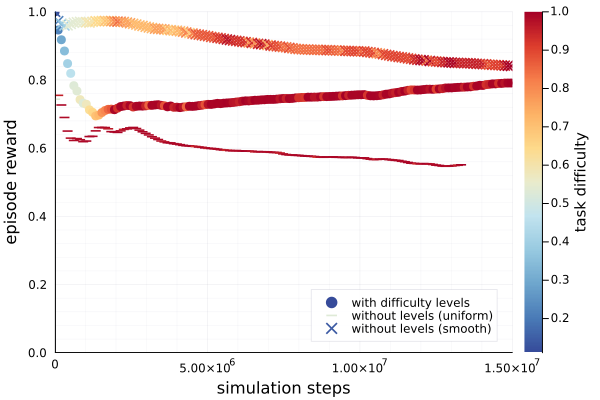} &
    \includegraphics[trim={0 0 3.1cm 0},clip,width=.275\textwidth]{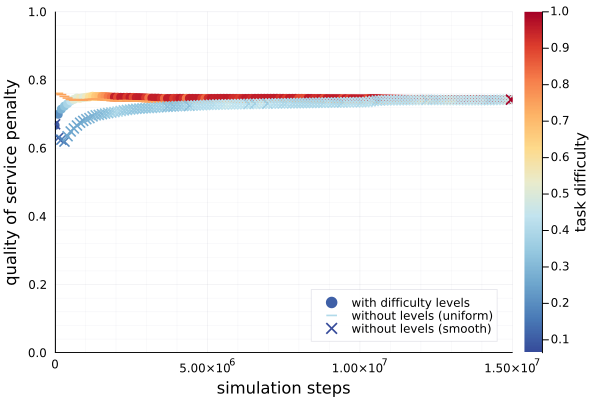} &
    \includegraphics[trim={0 0 0 0},clip,width=.325\textwidth]{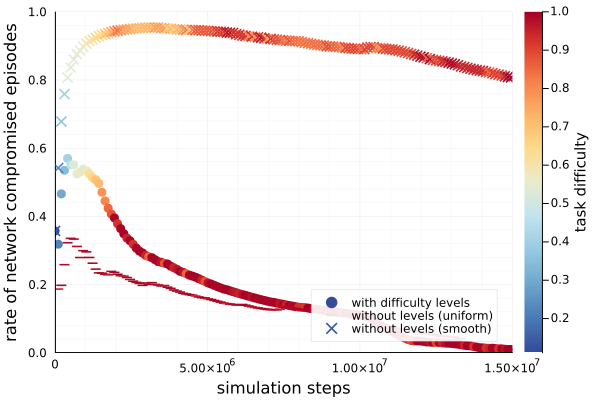}
\end{tabular}
\caption{Comparison of task selection strategies: one based on increasing difficulty levels, another selecting tasks uniformly at random, and a third using small variations. While 'smooth' task updates yield higher average rewards, the difficulty of tasks mastered is greater with the difficulty-based approach. Marker colors represent task difficulty during training, derived from hosts, subnets, and the variability of red and gray behaviors. Figure~\ref{fig:figsamplingstrategies_full} shows additional plots comparing these strategies across various network components.}
\label{fig:figsamplingstrategies}
\end{figure*}

As we discuss in Section \ref{subsec:task_selection}, during training, we select tasks
aiming to gradually increase the \emph{difficulty}  of a task. This section examines how this concrete strategy can result in a better performing policy.
For this, we compare the performance of the resulting policy guided by
difficulty with policies trained via alternative strategies.
Figure \ref{fig:figsamplingstrategies} illustrates the performance of the
strategies that we describe next.

\begin{description}
    \item[Difficulty driven] One of the policies is obtained via our
difficulty-driven task selection.
    \item[Uniformly random] A second policy is obtained presenting
diverse tasks (drawing from the same universe of tasks), but selecting their
parameters uniformly at random. 
    \item[Smooth changes] A third policy is obtained by varying tasks by a small
amount, but without being guided by difficulty. Concretely, training starts from
a task chosen arbitrarily, when the agent masters this task, the training task
is updated by varying one of several aspects by the minimum amount (e.g.,
increasing the number of hosts in a subnet by 1). To control
for the potential influence of the initial task in the this case, we train the
policy several times and report the average performance of all the policies
trained following the same approach. 
\end{description}

As we can see in Figure \ref{fig:figsamplingstrategies}, given a computing
budget, the difficulty-driven strategy masters harder tasks, with larger networks.
Sampling uniformly, tends to present an agent with tasks of medium difficulty,
while unguided smooth updates are unlikely to arrive at the highest complexity.
While there is more work needed to fully understand how to effectively select
tasks during training, our empirical results suggest that smooth updates with a
bias toward harder tasks could be a reasonable strategy, in the context of
training agents for network defense.

\subsection{Generalization}
\label{subsec:generalization}

Next, we evaluate the extent to which an agent trained with \emph{dynamic task
selection} is able to generalize to similar tasks, even if the agent never
experienced the task during training. Concretely, we compare the performance of
an agent trained to perform only one task (to mitigate a DoS attack) with
the performance of an agent that is trained experiencing a variety of tasks,
including the mitigation of an exfiltration and a ransomware attack.
However, the agent trained with dynamic task selection never experienced a
DoS attack (with a single flooding attacker).

In this part of our evaluation, each agent needs to defend against a DoS attack
in a network with 200 hosts during 100 episodes, while maintaining a minimum
quality of service. As we can see in Figure
\ref{fig:testing_performance}, the
policy trained with dynamic task selection outperforms the policy that was
trained to mitigate a DoS. This result suggests that, in addition to arriving at
a higher performance with shorter training, the diversity of tasks also helps to
arrive at more capable agents.

\begin{figure*}[t!]
    \centering
    \begin{tabular}{ccc}
        \includegraphics[trim={0 0 3.1cm 0},clip,width=.275\textwidth]{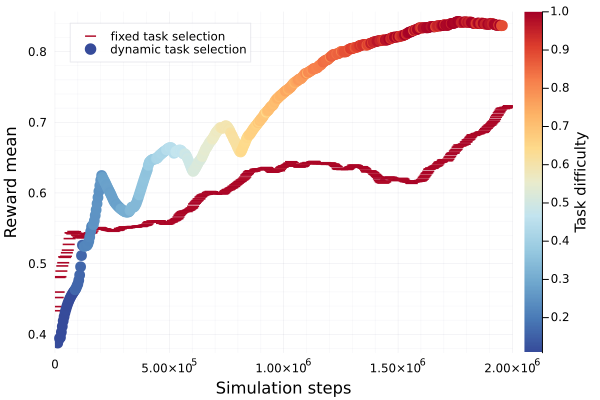} &
        \includegraphics[trim={0 0 3.1cm 0},clip,width=.275\textwidth]{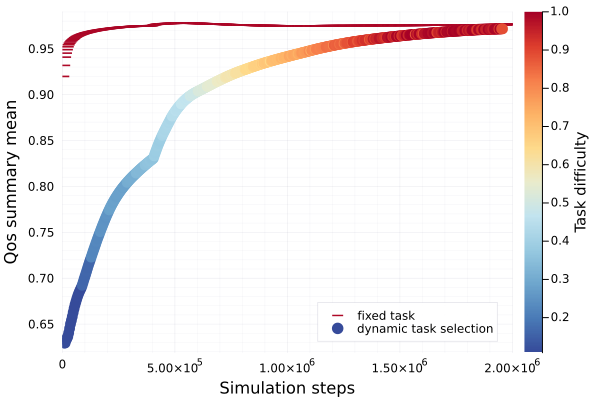} &
        \includegraphics[trim={0 0 0 0},clip,width=.325\textwidth]{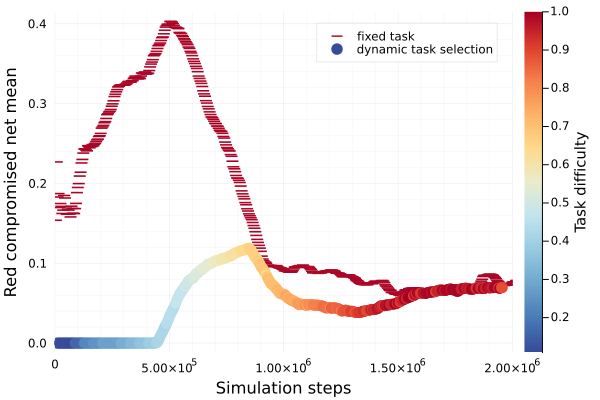} 
    \end{tabular}
\caption{Performance comparison of two policies: one trained with a fixed task to mitigate a traditional exfiltration attack, and the other trained with dynamic task selection, where the blue agent mitigates an exfiltration attack while the red agent interleaves gray-like actions between traditional red actions. Marker colors represent task difficulty during training, derived from hosts, subnets, and the variability of red and gray behaviors. Figure~\ref{fig:fig_morphing_full} shows additional plots comparing the two policies across other network components.}
\label{fig:fig_morphing}
\end{figure*}

Another form of generalization is related to obtaining policies that are robust
to variations in adversarial behavior with the intention to evade. We examine
the prospect of obtaining policies that are more robust against this type of
obfuscation via our approach. For this, we evaluate the performance of a policy
trained with dynamic task selection on a task similar to the one we describe in
\ref{subsec:scaling}. However, this time, the red agent gradually deviates from
its original behavior. Figure \ref{fig:fig_morphing} compares the performance of
two policies. As we can see, the performance during training is similar for
these two strategies. However, the agent training via dynamic task selection is
ultimately trained against a more sophisticated red TTP. 

\subsection{Curriculums with Goals}
\label{subsec:curriculus_with_goals}

\begin{figure*}[t!]
    \centering
    \begin{minipage}{.32\textwidth}
        \centering
        \includegraphics[width=\textwidth]{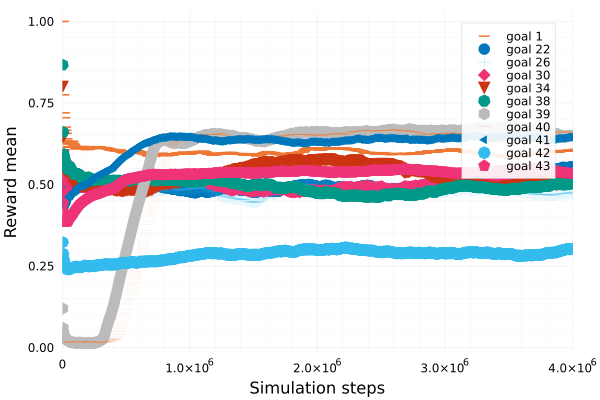}
        \caption{Mean reward for configurations, each fixing one of 11 goal-metric pairs.}
        \label{fig:11goals}
    \end{minipage}
    \hfill
    \begin{minipage}{.32\textwidth}
        \centering
        \includegraphics[width=\textwidth]{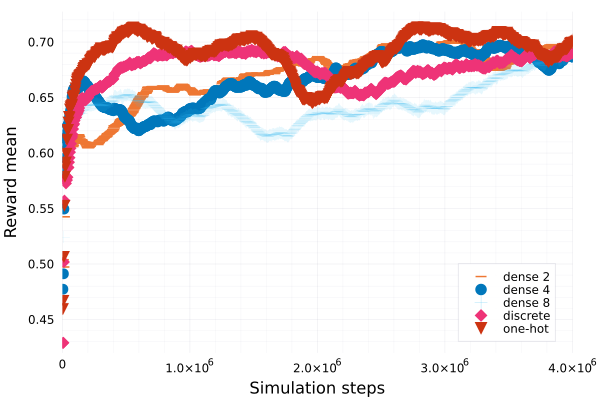}
        \caption{Mean reward for several different goal-metric pair representations.}
        \label{fig:goal_representation_reward}
    \end{minipage}
    \hfill
    \begin{minipage}{.32\textwidth}
        \centering
        \includegraphics[width=\textwidth]{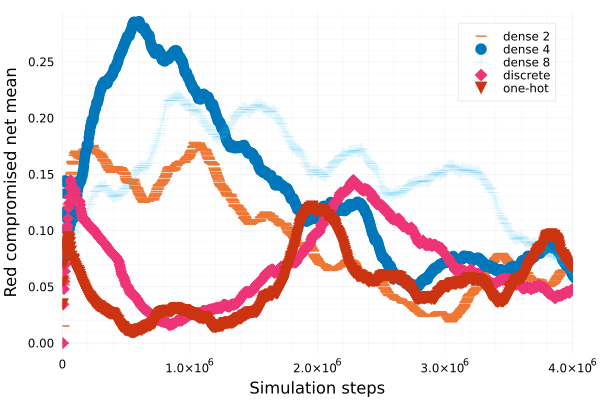}
        \caption{Fraction of episodes that result in a real compromise.}
        \label{fig:goal_representation_compromises}
    \end{minipage}
\end{figure*}

As we explained in Section \ref{subsec:goals}, our experiments include network
defense tasks where the agent must learn to master multiple tasks, corresponding
to 43 goal-metric pairs. While it is not easy to determine a total order of
these goal-metric pairs that corresponds to their difficulty, we compared the
difficulty of an agent to master tasks with each of the goal-metric pairs in a
subset of the 11 tasks that are more distinct, based on the goals and metrics.
In this comparison, all other parameters were sampled uniformly at random from
consistent sets. For example, the number of subnets in a task ranged from 2 to 5
and the number of hosts in each subnet ranged from 5 to 20. Figure
\ref{fig:11goals} shows the reward mean as the agent learns to master tasks in
each of the corresponding 11 sets. The goal-metric pairs included were: (1, 22,
26, 30, 34, 38, 39, 40, 41, 42, 43). For example, (goal, metric) pair 1 simply
requires the blue agent to learn to trigger any non-trivial action when there is
an active red agent and do nothing when a red agent is not active. There are no
other requirements about QoS or about minimizing unnecessary expensive actions.
Goal-metric pair 39 requires the blue agent to learn to use honey networks to
mitigate an active attack and to minimize the number of nontrivial blue actions.
Goal-metric pair 43 further requires the agent to learn to use honey networks to
convince the red agent that it has successfully compromised a real asset, and it
requires the agent to minimize the number of nontrivial blue actions and the
number of negative QoS events. The descriptions of all these goals are
included in Appendix \ref{appendix:pddl_terms}. 

Based on the difficulty to master these goals, we selected 6 that the agent was
able to reasonably master and used them in a curriculum to measure how five
different representations for goal-metric pairs resulted in higher performing
policies. The representations include a discrete representation, a one-hot
representation and three dense representations. For dense representations, we
use a trainable embedding Layer to transform discrete blue goals IDs into a
compact, dense vector representation. This embedding technique is enables the
treatment of discrete goal IDs within a high-dimensional continuous space.
Throughout the training process, back-propagation adjusts the parameters of the
goal embedding layer within our neural network to optimize the model's overall
performance.

Figures
\ref{fig:goal_representation_reward} and
\ref{fig:goal_representation_compromises} show that the one-hot representation
results in higher performance but the gains are not significant. The one-hot
representation drops entropy faster leading to more deterministic behavior.
However, further experimentation is needed to determine if a dense
representation would be better when the cardinality of the set of goal-metric
pairs is higher.

\subsection{Training}
\label{subsec:training}

In our experiments, we varied the following aspects of a task $T = (N, G)$. With
respect to the network dynamics in $N$, we varied the number of subnets in a
network, the number of hosts in each subnet, the amount of gray-agent diversity
and the amount of gray-traffic volume, and the specific TTP that drives the
behavior of the red agent. For the robustness experiments, we also varied the
probability that a red agent deviated from its original behavior, the deviation
diversity, and the deviation volume. The last two parameters are similar to the
amount and diversity of gray traffic. 

With respect to the initial conditions, we varied the number and identities of
the initially compromised hosts, as well as the placement of special host in the
network. Special hosts include those that have digital content that the
adversary seeks to exfiltrate and/or servers that the adversary seeks to compromise.

In this case, the goal balanced a set of QoS
and security subgoals. These were encoded in a sparse reward function that
reported the desirability of an outcome at the end of each episode. This
\emph{QoS-aware sparse reward} accounted for the following factors: The result
adds 1 when the blue agent wins and subtracts one when the blue agent loses.
Depending on the red TTP, the blue agent only wins if the red agent is deceived
and forced to compromise a fake target. This implies that preventing a
compromise only by isolating compromised hosts does not result in a blue win
with one exception. This \emph{QoS-aware sparse reward} accounts for the
possibility that there is no compromise (the red agent is inactive). In this
case, the best outcome is achieved by not disrupting QoS. QoS contributes a
number between -1 and 1 based on a normalized account of \emph{good QoS events}
and \emph{bad QoS events}. Additionally, it penalizes unnecessary blue actions
(host isolations and crown-jewel relocations). Penalizations contribute a number
in the interval [-1, 0] for each type of blue action. As we describe in
Appendix \ref{appendix1}, our implementation normalizes this reward function to
return values in the [0, 1] interval.

When the learning agent acts, it observes the following: The number of attempts
to relocate a protected host during an episode;  the number of attempts to
isolate a host during an episode; the number of isolated hosts during an
episode; the numeric ID of the last action; the number of log events associated
with a jewel search since the previous blue action; the number of log events
associated with passive discovery attempts since the previous blue action; the
number of log events associated with active discovery attempts since the
previous blue action; the number of failed HTTP connections reported by hosts
since the previous blue action; the number of failed SCP connections reported by
hosts since the previous blue action; the number of failed SSH connections
reported by hosts since the previous blue action; the number of successful HTTP
connections reported by hosts since the previous blue action; the number of
successful internal SCP connections reported by hosts since the previous blue
action; the number of successful SSH connections reported by hosts since the
previous blue action; the number of successful external SSH connections reported
by hosts since the previous blue action; and, a metric that summarizes QoS since
the previous blue action. 

\begin{figure*}[t!]
    \centering
    \begin{tabular}{ccc}
        \includegraphics[trim={0 0 3.1cm 0},clip,width=.275\textwidth]{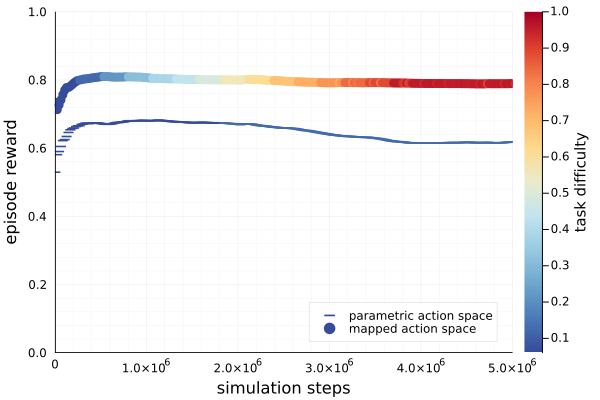} &
        \includegraphics[trim={0 0 3.1cm 0},clip,width=.275\textwidth]{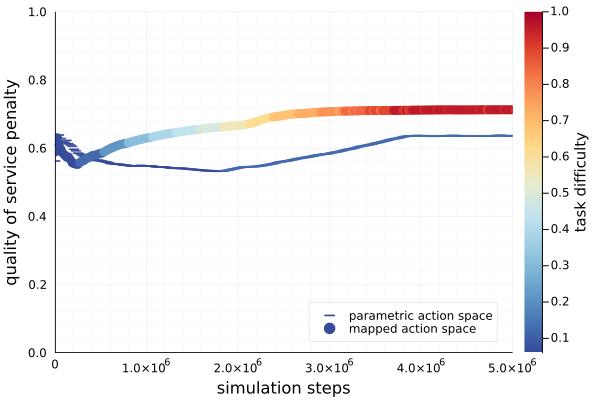} &
        \includegraphics[trim={0 0 0 0},clip,width=.325\textwidth]{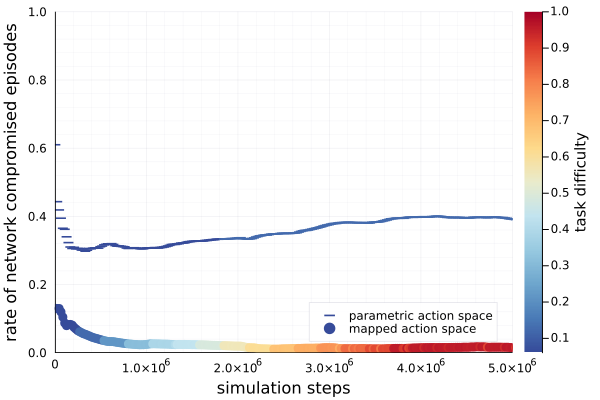} 
    \end{tabular}
\caption{Comparison of two learning strategies that differ only in the available actions: one uses action representations, while the other allows the agent to select actions based on the target host. Learning stagnates with a large action space. In both strategies, observation spaces, reward functions, and network dynamics are the same. Figure~\ref{fig:action_representations_full} shows additional plots comparing these strategies across other network components.}
\label{fig:action_representations}
\end{figure*}

Depending on the task, the blue agent needed to mitigate an exfiltration, a
DDoS, a DoS, or a ransomware attack by performing one of the following actions:
isolating a protected host, isolating a suspicious host, relocating a protected
host, reimaging a suspicious host, or migrating a suspicious host to a honey
network. Using action representations~\cite{chandak_learning_2019}, we map a
small number of actions to a larger set of parametric actions where the subject
host of the action needs to be specified. This strategy already proves some
learning benefits, as we illustrate in Figure \ref{fig:action_representations}.
This plot illustrates that, given two training settings that only differ in the
action spaces, when the action space is too large, learning is more difficult.

\subsection{Algorithm and Hyperparameters}
\label{subsection:algorithm}

We use the Proximal Policy Optimization (PPO) implementation by RLlib with the
hyperparamters in Table \ref{tab:ppo_hyperparameters}. We selected these
hyperparameters through a combination of experimentation and recommendations by
Andrychowicz et al.~\cite{andrychowicz_what_2021}. For example, we use generalized
advantage estimation (GAE); we applied observation normalization to the
observation space; and we used separate value and policy networks. To min-max
scale the observation features across tasks in the task space, we scaled
the minimum and maximum bounds accordingly with the number of hosts and the
horizon of the episode. This ensures the observation space remains consistent as
we discuss in Section \ref{states_and_actions}. 

\begin{table}[htb]
    \tiny
    \caption{PPO Agent Training Hyperparameters}
    \centering
    \begin{tabular}{|c|c|}
    \hline
    \textbf{Hyperparameter} & \textbf{Value} \\ \hline \hline
    gamma & 0.99 \\ \hline
    lr & 0.0001 \\ \hline
    fcnet\_activation & ReLU  \\ \hline
    use\_critic &  True \\ \hline
    use\_gae &  True  \\ \hline
    lambda\_ (\textbf{GAE lambda parameter}) &  1.0 \\ \hline
    vf\_share\_layers &  False  \\ \hline
    fcnet\_hiddens (\textbf{policy}) &  [128, 128, 128, 128] \\ \hline
    fcnet\_hiddens (\textbf{value}) &  [256, 256, 256, 256] \\ \hline
    kl\_coeff &  0.5 \\ \hline
    kl\_target & 0.01 \\ \hline
    train\_batch\_size & 4000  \\ \hline
    sgd\_minibatch\_size & 128  \\ \hline
    num\_sgd\_iter & 30 \\ \hline
    shuffle\_sequences & True  \\ \hline
    vf\_loss\_coeff & 1.0 \\ \hline
    entropy\_coeff & 0.005 \\ \hline
    entropy\_coeff\_schedule & None \\ \hline
    clip\_param & 0.3 \\ \hline
    vf\_clip\_param & 1.0 \\ \hline
    grad\_clip & 500 \\ \hline
    evaluation\_duration\_unit & episodes \\ \hline
    evaluation\_duration & 20 \\ \hline
    evaluation\_config\_explore & True \\ \hline
    \end{tabular}
    \label{tab:ppo_hyperparameters}
\end{table}

Dynamic task selection can also work in combination with other algorithms, such
soft actor critic algorithms (i.e., SAC) and DQN.  Figure
\ref{fig:fig_algorithms}, in the Appendix, compares the performance for a
universe with tasks that vary network dynamics but keep the goal-pair constant.

\section{Discussion and Conclusion}

This paper studies the role of training network defenders with task universes,
as opposed to training with a single task. We highlight scalability and
generalizability as some of the benefits of training and evaluating
cybersecurity agents via the specification of comprehensive sets of tasks, rather
than small sets of ``representative'' tasks. Furthermore, by presenting a broad
set of tasks during training, it is possible to train agents to mitigate
realistic threats. This represents an important milestone toward being able to
delegate useful cybersecurity tasks to an autonomous agent. 

However, while we show that training with dynamic task selection can result in
shorter training and higher generalizability, more work is needed to identify
optimal task selection strategies during training. Similarly, further research
must address the problem of defining adequate task selection strategies to
evaluate policies in adversarial settings. Moreover, we believe that building
protection mechanisms for RL-enabled decisions requires more than just training
an agent with a broader task universe. In future work, we are investigating how
dynamic task selection can be combined with RL algorithms that integrate
probabilistic inference to infer relations between the current task and other
tasks that the agent has already mastered. We hypothesize that this can help to
minimize some forms of adversarial manipulation where an adversary obfuscates a
well-known TTP to avoid detection. 





\section*{Acknowledgments}

This paper builds upon FARLAND~\cite{molina-markham_network_2021}, a tool developed through the collaborative efforts of researchers at MITRE and the NSA. We gratefully acknowledge the contributions of the authors of this paper, as well as the many developers and reviewers whose expertise and dedication have shaped FARLAND into what it is today.

Source code to replicate our results and build upon our approach is forthcoming.


\bibliographystyle{ACM-Reference-Format}
\bibliography{auto}


\begin{thebibliography}{45}


\ifx \showCODEN    \undefined \def \showCODEN     #1{\unskip}     \fi
\ifx \showDOI      \undefined \def \showDOI       #1{#1}\fi
\ifx \showISBNx    \undefined \def \showISBNx     #1{\unskip}     \fi
\ifx \showISBNxiii \undefined \def \showISBNxiii  #1{\unskip}     \fi
\ifx \showISSN     \undefined \def \showISSN      #1{\unskip}     \fi
\ifx \showLCCN     \undefined \def \showLCCN      #1{\unskip}     \fi
\ifx \shownote     \undefined \def \shownote      #1{#1}          \fi
\ifx \showarticletitle \undefined \def \showarticletitle #1{#1}   \fi
\ifx \showURL      \undefined \def \showURL       {\relax}        \fi
\providecommand\bibfield[2]{#2}
\providecommand\bibinfo[2]{#2}
\providecommand\natexlab[1]{#1}
\providecommand\showeprint[2][]{arXiv:#2}

\bibitem[\protect\citeauthoryear{CodaLab Competitions}{cag}{2024}]%
        {cage_challenge}
 \bibinfo{year}{2024}\natexlab{}.
\newblock \bibinfo{title}{{C}{A}{G}{E} Challenge 4}.
\newblock   (\bibinfo{year}{2024}).
\newblock
\showURL{%
\url{https://codalab.lisn.upsaclay.fr/competitions/17672#results}}


\bibitem[\protect\citeauthoryear{Andrew, Spillard, Collyer, and Dhir}{Andrew
  et~al\mbox{.}}{2022}]%
        {andrew_developing_2022}
\bibfield{author}{\bibinfo{person}{Alex Andrew}, \bibinfo{person}{Sam
  Spillard}, \bibinfo{person}{Joshua Collyer}, {and} \bibinfo{person}{Neil
  Dhir}.} \bibinfo{year}{2022}\natexlab{}.
\newblock \showarticletitle{Developing {Optimal} {Causal} {Cyber}-{Defence}
  {Agents} via {Cyber} {Security} {Simulation}}. In \bibinfo{booktitle}{{\em
  Workshop on {Machine} {Learning} for {Cybersecurity} ({ML4Cyber})}}.
\newblock


\bibitem[\protect\citeauthoryear{Andrychowicz, Raichuk, Stańczyk, Orsini,
  Girgin, Marinier, Hussenot, Geist, Pietquin, Michalski, Gelly, and
  Bachem}{Andrychowicz et~al\mbox{.}}{2021}]%
        {andrychowicz_what_2021}
\bibfield{author}{\bibinfo{person}{Marcin Andrychowicz}, \bibinfo{person}{Anton
  Raichuk}, \bibinfo{person}{Piotr Stańczyk}, \bibinfo{person}{Manu Orsini},
  \bibinfo{person}{Sertan Girgin}, \bibinfo{person}{Raphaël Marinier},
  \bibinfo{person}{Leonard Hussenot}, \bibinfo{person}{Matthieu Geist},
  \bibinfo{person}{Olivier Pietquin}, \bibinfo{person}{Marcin Michalski},
  \bibinfo{person}{Sylvain Gelly}, {and} \bibinfo{person}{Olivier Bachem}.}
  \bibinfo{year}{2021}\natexlab{}.
\newblock \showarticletitle{What {Matters} for {On}-{Policy} {Deep}
  {Actor}-{Critic} {Methods}? {A} {Large}-{Scale} {Study}}. In
  \bibinfo{booktitle}{{\em International {Conference} on {Learning}
  {Representations}}}.
\newblock
\showURL{%
\url{https://openreview.net/forum?id=nIAxjsniDzg}}


\bibitem[\protect\citeauthoryear{Applebaum, Dennler, Dwyer, Moskowitz, Nguyen,
  Nichols, Park, Rachwalski, Rau, Webster, and Wolk}{Applebaum
  et~al\mbox{.}}{2022}]%
        {applebaum_bridging_2022}
\bibfield{author}{\bibinfo{person}{Andy Applebaum}, \bibinfo{person}{Camron
  Dennler}, \bibinfo{person}{Patrick Dwyer}, \bibinfo{person}{Marina
  Moskowitz}, \bibinfo{person}{Harold Nguyen}, \bibinfo{person}{Nicole
  Nichols}, \bibinfo{person}{Nicole Park}, \bibinfo{person}{Paul Rachwalski},
  \bibinfo{person}{Frank Rau}, \bibinfo{person}{Adrian Webster}, {and}
  \bibinfo{person}{Melody Wolk}.} \bibinfo{year}{2022}\natexlab{}.
\newblock \showarticletitle{Bridging {Automated} to {Autonomous} {Cyber}
  {Defense}: {Foundational} {Analysis} of {Tabular} {Q}-{Learning}}. In
  \bibinfo{booktitle}{{\em Proceedings of the 15th {ACM} {Workshop} on
  {Artificial} {Intelligence} and {Security}}}. \bibinfo{publisher}{ACM},
  \bibinfo{address}{Los Angeles CA USA}, \bibinfo{pages}{149--159}.
\newblock
\showISBNx{978-1-4503-9880-0}
\showDOI{%
\url{https://doi.org/10.1145/3560830.3563732}}


\bibitem[\protect\citeauthoryear{Baillie, Standen, Schwartz, Docking, Bowman,
  and Kim}{Baillie et~al\mbox{.}}{2020}]%
        {baillie_cyborg_2020}
\bibfield{author}{\bibinfo{person}{Callum Baillie}, \bibinfo{person}{Maxwell
  Standen}, \bibinfo{person}{Jonathon Schwartz}, \bibinfo{person}{Michael
  Docking}, \bibinfo{person}{David Bowman}, {and} \bibinfo{person}{Junae Kim}.}
  \bibinfo{year}{2020}\natexlab{}.
\newblock \showarticletitle{Cyborg: {An} autonomous cyber operations research
  gym}.
\newblock \bibinfo{journal}{{\em arXiv preprint arXiv:2002.10667\/}}
  (\bibinfo{year}{2020}).
\newblock


\bibitem[\protect\citeauthoryear{Balduzzi, Garnelo, Bachrach, Czarnecki,
  Perolat, Jaderberg, and Graepel}{Balduzzi et~al\mbox{.}}{2019}]%
        {balduzzi_open-ended_2019}
\bibfield{author}{\bibinfo{person}{David Balduzzi}, \bibinfo{person}{Marta
  Garnelo}, \bibinfo{person}{Yoram Bachrach}, \bibinfo{person}{Wojciech
  Czarnecki}, \bibinfo{person}{Julien Perolat}, \bibinfo{person}{Max
  Jaderberg}, {and} \bibinfo{person}{Thore Graepel}.}
  \bibinfo{year}{2019}\natexlab{}.
\newblock \showarticletitle{Open-ended learning in symmetric zero-sum games}.
  In \bibinfo{booktitle}{{\em International {Conference} on {Machine}
  {Learning}}}. \bibinfo{publisher}{PMLR}, \bibinfo{pages}{434--443}.
\newblock


\bibitem[\protect\citeauthoryear{Basori and Malebary}{Basori and
  Malebary}{2020}]%
        {basori_deep_2020}
\bibfield{author}{\bibinfo{person}{Ahmad~Hoirul Basori} {and}
  \bibinfo{person}{Sharaf~Jameel Malebary}.} \bibinfo{year}{2020}\natexlab{}.
\newblock \showarticletitle{Deep {Reinforcement} {Learning} for {Adaptive}
  {Cyber} {Defense} and {Attacker}'s {Pattern} {Identification}}.
\newblock In \bibinfo{booktitle}{{\em Advances in {Cyber} {Security}
  {Analytics} and {Decision} {Systems}}}. \bibinfo{publisher}{Springer},
  \bibinfo{pages}{15--26}.
\newblock


\bibitem[\protect\citeauthoryear{Bates, Mavroudis, and Hicks}{Bates
  et~al\mbox{.}}{2023}]%
        {bates_reward_2023}
\bibfield{author}{\bibinfo{person}{Elizabeth Bates}, \bibinfo{person}{Vasilios
  Mavroudis}, {and} \bibinfo{person}{Chris Hicks}.}
  \bibinfo{year}{2023}\natexlab{}.
\newblock \showarticletitle{Reward {Shaping} for {Happier} {Autonomous} {Cyber}
  {Security} {Agents}}. In \bibinfo{booktitle}{{\em Proceedings of the 16th
  {ACM} {Workshop} on {Artificial} {Intelligence} and {Security}}}.
  \bibinfo{publisher}{ACM}, \bibinfo{address}{Copenhagen Denmark},
  \bibinfo{pages}{221--232}.
\newblock
\showISBNx{9798400702600}
\showDOI{%
\url{https://doi.org/10.1145/3605764.3623916}}


\bibitem[\protect\citeauthoryear{Beltiukov, Guo, Gupta, and
  Willinger}{Beltiukov et~al\mbox{.}}{2023}]%
        {beltiukov_search_2023}
\bibfield{author}{\bibinfo{person}{Roman Beltiukov}, \bibinfo{person}{Wenbo
  Guo}, \bibinfo{person}{Arpit Gupta}, {and} \bibinfo{person}{Walter
  Willinger}.} \bibinfo{year}{2023}\natexlab{}.
\newblock \showarticletitle{In {Search} of {NetUnicorn}: {A}
  {Data}-{Collection} {Platform} to {Develop} {Generalizable} {ML} {Models} for
  {Network} {Security} {Problems}}. In \bibinfo{booktitle}{{\em Proceedings of
  the 2023 {ACM} {SIGSAC} {Conference} on {Computer} and {Communications}
  {Security}}} {\em (\bibinfo{series}{{CCS} '23})}.
  \bibinfo{publisher}{Association for Computing Machinery},
  \bibinfo{address}{New York, NY, USA}, \bibinfo{pages}{2217--2231}.
\newblock
\showISBNx{9798400700507}
\showDOI{%
\url{https://doi.org/10.1145/3576915.3623075}}


\bibitem[\protect\citeauthoryear{Bengio, Louradour, Collobert, and
  Weston}{Bengio et~al\mbox{.}}{2009}]%
        {bengio_curriculum_2009}
\bibfield{author}{\bibinfo{person}{Yoshua Bengio}, \bibinfo{person}{Jérôme
  Louradour}, \bibinfo{person}{Ronan Collobert}, {and} \bibinfo{person}{Jason
  Weston}.} \bibinfo{year}{2009}\natexlab{}.
\newblock \showarticletitle{Curriculum learning}. In \bibinfo{booktitle}{{\em
  {ICML} '09}}.
\newblock


\bibitem[\protect\citeauthoryear{Chandak, Theocharous, Kostas, Jordan, and
  Thomas}{Chandak et~al\mbox{.}}{2019}]%
        {chandak_learning_2019}
\bibfield{author}{\bibinfo{person}{Yash Chandak}, \bibinfo{person}{Georgios
  Theocharous}, \bibinfo{person}{James Kostas}, \bibinfo{person}{Scott Jordan},
  {and} \bibinfo{person}{Philip Thomas}.} \bibinfo{year}{2019}\natexlab{}.
\newblock \showarticletitle{Learning {Action} {Representations} for
  {Reinforcement} {Learning}}. In \bibinfo{booktitle}{{\em Proceedings of the
  36th {International} {Conference} on {Machine} {Learning}}} {\em
  (\bibinfo{series}{Proceedings of {Machine} {Learning} {Research}})},
  \bibfield{editor}{\bibinfo{person}{Kamalika Chaudhuri} {and}
  \bibinfo{person}{Ruslan Salakhutdinov}} (Eds.), Vol.~\bibinfo{volume}{97}.
  \bibinfo{publisher}{PMLR}, \bibinfo{pages}{941--950}.
\newblock
\showURL{%
\url{https://proceedings.mlr.press/v97/chandak19a.html}}


\bibitem[\protect\citeauthoryear{Doncieux, Filliat, Díaz-Rodríguez,
  Hospedales, Duro, Coninx, Roijers, Girard, Perrin, and Sigaud}{Doncieux
  et~al\mbox{.}}{2018}]%
        {doncieux_open-ended_2018}
\bibfield{author}{\bibinfo{person}{Stephane Doncieux}, \bibinfo{person}{David
  Filliat}, \bibinfo{person}{Natalia Díaz-Rodríguez},
  \bibinfo{person}{Timothy Hospedales}, \bibinfo{person}{Richard Duro},
  \bibinfo{person}{Alexandre Coninx}, \bibinfo{person}{Diederik~M Roijers},
  \bibinfo{person}{Benoît Girard}, \bibinfo{person}{Nicolas Perrin}, {and}
  \bibinfo{person}{Olivier Sigaud}.} \bibinfo{year}{2018}\natexlab{}.
\newblock \showarticletitle{Open-ended learning: a conceptual framework based
  on representational redescription}.
\newblock \bibinfo{journal}{{\em Frontiers in neurorobotics\/}}
  \bibinfo{volume}{12} (\bibinfo{year}{2018}), \bibinfo{pages}{59}.
\newblock


\bibitem[\protect\citeauthoryear{Duan, Schulman, Chen, Bartlett, Sutskever, and
  Abbeel}{Duan et~al\mbox{.}}{2016}]%
        {duan_rl2_2016}
\bibfield{author}{\bibinfo{person}{Yan Duan}, \bibinfo{person}{John Schulman},
  \bibinfo{person}{Xi Chen}, \bibinfo{person}{Peter~L. Bartlett},
  \bibinfo{person}{Ilya Sutskever}, {and} \bibinfo{person}{Pieter Abbeel}.}
  \bibinfo{year}{2016}\natexlab{}.
\newblock \bibinfo{title}{{RL}{\textbackslash}ˆ2{\textbackslash}: {Fast}
  {Reinforcement} {Learning} via {Slow} {Reinforcement} {Learning}}.
\newblock   (\bibinfo{year}{2016}).
\newblock
\showDOI{%
\url{https://doi.org/10.48550/ARXIV.1611.02779}}


\bibitem[\protect\citeauthoryear{Eysenbach, Gupta, Ibarz, and Levine}{Eysenbach
  et~al\mbox{.}}{2018}]%
        {eysenbach_diversity_2018}
\bibfield{author}{\bibinfo{person}{Benjamin Eysenbach},
  \bibinfo{person}{Abhishek Gupta}, \bibinfo{person}{Julian Ibarz}, {and}
  \bibinfo{person}{Sergey Levine}.} \bibinfo{year}{2018}\natexlab{}.
\newblock \bibinfo{title}{Diversity is {All} {You} {Need}: {Learning} {Skills}
  without a {Reward} {Function}}.
\newblock   (\bibinfo{year}{2018}).
\newblock
\showDOI{%
\url{https://doi.org/10.48550/ARXIV.1802.06070}}


\bibitem[\protect\citeauthoryear{Feng and Xu}{Feng and Xu}{2017}]%
        {feng_deep_2017}
\bibfield{author}{\bibinfo{person}{Ming Feng} {and} \bibinfo{person}{Hao Xu}.}
  \bibinfo{year}{2017}\natexlab{}.
\newblock \showarticletitle{Deep reinforecement learning based optimal defense
  for cyber-physical system in presence of unknown cyber-attack}. In
  \bibinfo{booktitle}{{\em 2017 {IEEE} {Symposium} {Series} on {Computational}
  {Intelligence} ({SSCI})}}. \bibinfo{publisher}{IEEE},
  \bibinfo{address}{Honolulu, HI}, \bibinfo{pages}{1--8}.
\newblock
\showISBNx{978-1-5386-2726-6}
\showDOI{%
\url{https://doi.org/10.1109/SSCI.2017.8285298}}


\bibitem[\protect\citeauthoryear{Finn, Rajeswaran, Kakade, and Levine}{Finn
  et~al\mbox{.}}{2019}]%
        {finn_online_2019}
\bibfield{author}{\bibinfo{person}{Chelsea Finn}, \bibinfo{person}{Aravind
  Rajeswaran}, \bibinfo{person}{Sham Kakade}, {and} \bibinfo{person}{Sergey
  Levine}.} \bibinfo{year}{2019}\natexlab{}.
\newblock \bibinfo{title}{Online {Meta}-{Learning}}.
\newblock   (\bibinfo{year}{2019}).
\newblock
\showDOI{%
\url{https://doi.org/10.48550/ARXIV.1902.08438}}


\bibitem[\protect\citeauthoryear{Foley, Hicks, Highnam, and Mavroudis}{Foley
  et~al\mbox{.}}{2022}]%
        {foley_autonomous_2022}
\bibfield{author}{\bibinfo{person}{Myles Foley}, \bibinfo{person}{Chris Hicks},
  \bibinfo{person}{Kate Highnam}, {and} \bibinfo{person}{Vasilios Mavroudis}.}
  \bibinfo{year}{2022}\natexlab{}.
\newblock \showarticletitle{Autonomous {Network} {Defence} using
  {Reinforcement} {Learning}}. In \bibinfo{booktitle}{{\em Proceedings of the
  2022 {ACM} on {Asia} {Conference} on {Computer} and {Communications}
  {Security}}}. \bibinfo{pages}{1252--1254}.
\newblock


\bibitem[\protect\citeauthoryear{Gangupantulu, Cody, Park, Rahman, Eisenbeiser,
  Radke, Clark, and Redino}{Gangupantulu et~al\mbox{.}}{2022}]%
        {gangupantulu_using_2022}
\bibfield{author}{\bibinfo{person}{Rohit Gangupantulu}, \bibinfo{person}{Tyler
  Cody}, \bibinfo{person}{Paul Park}, \bibinfo{person}{Abdul Rahman},
  \bibinfo{person}{Logan Eisenbeiser}, \bibinfo{person}{Dan Radke},
  \bibinfo{person}{Ryan Clark}, {and} \bibinfo{person}{Christopher Redino}.}
  \bibinfo{year}{2022}\natexlab{}.
\newblock \showarticletitle{Using cyber terrain in reinforcement learning for
  penetration testing}. In \bibinfo{booktitle}{{\em 2022 {IEEE} {International}
  {Conference} on {Omni}-layer {Intelligent} {Systems} ({COINS})}}.
  \bibinfo{publisher}{IEEE}, \bibinfo{pages}{1--8}.
\newblock


\bibitem[\protect\citeauthoryear{Hammar and Stadler}{Hammar and
  Stadler}{2024}]%
        {hammar_learning_2024}
\bibfield{author}{\bibinfo{person}{Kim Hammar} {and} \bibinfo{person}{Rolf
  Stadler}.} \bibinfo{year}{2024}\natexlab{}.
\newblock \showarticletitle{Learning {Near}-{Optimal} {Intrusion} {Responses}
  {Against} {Dynamic} {Attackers}}.
\newblock \bibinfo{journal}{{\em IEEE Transactions on Network and Service
  Management\/}} \bibinfo{volume}{21}, \bibinfo{number}{1}
  (\bibinfo{date}{Feb.} \bibinfo{year}{2024}), \bibinfo{pages}{1158--1177}.
\newblock
\showISSN{1932-4537, 2373-7379}
\showDOI{%
\url{https://doi.org/10.1109/TNSM.2023.3293413}}


\bibitem[\protect\citeauthoryear{Havens, Jiang, and Sarkar}{Havens
  et~al\mbox{.}}{[n. d.]}]%
        {havens_online_nodate}
\bibfield{author}{\bibinfo{person}{Aaron Havens}, \bibinfo{person}{Zhanhong
  Jiang}, {and} \bibinfo{person}{Soumik Sarkar}.} \bibinfo{year}{[n.
  d.]}\natexlab{}.
\newblock \showarticletitle{Online {Robust} {Policy} {Learning} in the
  {Presence} of {Unknown} {Adversaries}}.
\newblock  (\bibinfo{year}{[n. d.]}).
\newblock


\bibitem[\protect\citeauthoryear{Jacobs, Beltiukov, Willinger, Ferreira, Gupta,
  and Granville}{Jacobs et~al\mbox{.}}{2022}]%
        {jacobs_aiml_2022}
\bibfield{author}{\bibinfo{person}{A.~S. Jacobs}, \bibinfo{person}{R.
  Beltiukov}, \bibinfo{person}{W. Willinger}, \bibinfo{person}{R.~A. Ferreira},
  \bibinfo{person}{A. Gupta}, {and} \bibinfo{person}{L.~Z. Granville}.}
  \bibinfo{year}{2022}\natexlab{}.
\newblock \showarticletitle{{AI}/{ML} and {Network} {Security}: {The} {Emperor}
  has no {Clothes}}. In \bibinfo{booktitle}{{\em Proceedings of the 2022 {ACM}
  {SIGSAC} {Conference} on {Computer} and {Communications} {Security}}} {\em
  (\bibinfo{series}{{CCS} '22})}. \bibinfo{publisher}{Association for Computing
  Machinery}, \bibinfo{address}{New York, NY, USA}.
\newblock


\bibitem[\protect\citeauthoryear{Janisch, Pevný, and Lisý}{Janisch
  et~al\mbox{.}}{2023}]%
        {janisch_nasimemu_2023}
\bibfield{author}{\bibinfo{person}{Jaromír Janisch}, \bibinfo{person}{Tomáš
  Pevný}, {and} \bibinfo{person}{Viliam Lisý}.}
  \bibinfo{year}{2023}\natexlab{}.
\newblock \bibinfo{title}{{NASimEmu}: {Network} {Attack} {Simulator} \&
  {Emulator} for {Training} {Agents} {Generalizing} to {Novel} {Scenarios}}.
\newblock   (\bibinfo{date}{Aug.} \bibinfo{year}{2023}).
\newblock
\showURL{%
\url{http://arxiv.org/abs/2305.17246}}
\newblock
\shownote{arXiv:2305.17246 [cs].}


\bibitem[\protect\citeauthoryear{Kalashnikov, Varley, Chebotar, Swanson,
  Jonschkowski, Finn, Levine, and Hausman}{Kalashnikov et~al\mbox{.}}{2021}]%
        {kalashnikov_mt-opt_2021}
\bibfield{author}{\bibinfo{person}{Dmitry Kalashnikov}, \bibinfo{person}{Jacob
  Varley}, \bibinfo{person}{Yevgen Chebotar}, \bibinfo{person}{Benjamin
  Swanson}, \bibinfo{person}{Rico Jonschkowski}, \bibinfo{person}{Chelsea
  Finn}, \bibinfo{person}{Sergey Levine}, {and} \bibinfo{person}{Karol
  Hausman}.} \bibinfo{year}{2021}\natexlab{}.
\newblock \bibinfo{title}{{MT}-{Opt}: {Continuous} {Multi}-{Task} {Robotic}
  {Reinforcement} {Learning} at {Scale}}.
\newblock   (\bibinfo{year}{2021}).
\newblock
\showDOI{%
\url{https://doi.org/10.48550/ARXIV.2104.08212}}


\bibitem[\protect\citeauthoryear{Kanagawa and Kaneko}{Kanagawa and
  Kaneko}{2019}]%
        {kanagawa_rogue-gym_2019}
\bibfield{author}{\bibinfo{person}{Yuji Kanagawa} {and}
  \bibinfo{person}{Tomoyuki Kaneko}.} \bibinfo{year}{2019}\natexlab{}.
\newblock \showarticletitle{Rogue-gym: {A} new challenge for generalization in
  reinforcement learning}. In \bibinfo{booktitle}{{\em 2019 {IEEE} {Conference}
  on {Games} ({CoG})}}. \bibinfo{publisher}{IEEE}, \bibinfo{pages}{1--8}.
\newblock


\bibitem[\protect\citeauthoryear{Li, El~Rami, Taylor, Rao, and Kunz}{Li
  et~al\mbox{.}}{2022}]%
        {li_cygil_2022}
\bibfield{author}{\bibinfo{person}{Li Li}, \bibinfo{person}{Jean-Pierre~S.
  El~Rami}, \bibinfo{person}{Adrian Taylor}, \bibinfo{person}{James~Hailing
  Rao}, {and} \bibinfo{person}{Thomas Kunz}.} \bibinfo{year}{2022}\natexlab{}.
\newblock \showarticletitle{Enabling A Network {A}{I} Gym for Autonomous Cyber
  Agents}. In \bibinfo{booktitle}{{\em 2022 {IEEE} {International Conference}
  on {Computational Science} and {Computational Intelligence} ({CSCI})}}.
  \bibinfo{publisher}{IEEE}, \bibinfo{pages}{172--177}.
\newblock


\bibitem[\protect\citeauthoryear{Li, Fayad, and Taylor}{Li
  et~al\mbox{.}}{2021}]%
        {li_cygil_2021}
\bibfield{author}{\bibinfo{person}{Li Li}, \bibinfo{person}{Raed Fayad}, {and}
  \bibinfo{person}{Adrian Taylor}.} \bibinfo{year}{2021}\natexlab{}.
\newblock \showarticletitle{Cygil: {A} cyber gym for training autonomous agents
  over emulated network systems}.
\newblock \bibinfo{journal}{{\em arXiv preprint arXiv:2109.03331\/}}
  (\bibinfo{year}{2021}).
\newblock


\bibitem[\protect\citeauthoryear{Liu, Jia, Wen, Hu, Chen, Fan, Hu, and
  Yang}{Liu et~al\mbox{.}}{2021}]%
        {liu_towards_2021}
\bibfield{author}{\bibinfo{person}{Xiangyu Liu}, \bibinfo{person}{Hangtian
  Jia}, \bibinfo{person}{Ying Wen}, \bibinfo{person}{Yujing Hu},
  \bibinfo{person}{Yingfeng Chen}, \bibinfo{person}{Changjie Fan},
  \bibinfo{person}{Zhipeng Hu}, {and} \bibinfo{person}{Yaodong Yang}.}
  \bibinfo{year}{2021}\natexlab{}.
\newblock \showarticletitle{Towards unifying behavioral and response diversity
  for open-ended learning in zero-sum games}.
\newblock \bibinfo{journal}{{\em Advances in Neural Information Processing
  Systems\/}}  \bibinfo{volume}{34} (\bibinfo{year}{2021}),
  \bibinfo{pages}{941--952}.
\newblock


\bibitem[\protect\citeauthoryear{McCarthy}{McCarthy}{2023}]%
        {mccarthy_primaite_2023}
\bibfield{author}{\bibinfo{person}{Chris McCarthy}.}
  \bibinfo{year}{2023}\natexlab{}.
\newblock \bibinfo{title}{Primaite}.
\newblock   (\bibinfo{year}{2023}).
\newblock
\showURL{%
\url{https://github.com/Autonomous-Resilient-Cyber-Defence/PrimAITE}}


\bibitem[\protect\citeauthoryear{Molina-Markham, Winder, and
  Ridley}{Molina-Markham et~al\mbox{.}}{2021}]%
        {molina-markham_network_2021}
\bibfield{author}{\bibinfo{person}{Andres Molina-Markham},
  \bibinfo{person}{Ransom~K. Winder}, {and} \bibinfo{person}{Ahmad Ridley}.}
  \bibinfo{year}{2021}\natexlab{}.
\newblock \showarticletitle{Network {Defense} is {Not} a {Game}}.
\newblock \bibinfo{journal}{{\em CoRR\/}}  \bibinfo{volume}{abs/2104.10262}
  (\bibinfo{year}{2021}).
\newblock
\showURL{%
\url{https://arxiv.org/abs/2104.10262}}


\bibitem[\protect\citeauthoryear{Nyberg and Johnson}{Nyberg and
  Johnson}{2023}]%
        {nyberg_training_2023}
\bibfield{author}{\bibinfo{person}{Jakob Nyberg} {and} \bibinfo{person}{Pontus
  Johnson}.} \bibinfo{year}{2023}\natexlab{}.
\newblock \bibinfo{title}{Training {Automated} {Defense} {Strategies} {Using}
  {Graph}-based {Cyber} {Attack} {Simulations}}.
\newblock   (\bibinfo{date}{April} \bibinfo{year}{2023}).
\newblock
\showURL{%
\url{http://arxiv.org/abs/2304.11084}}
\newblock
\shownote{arXiv:2304.11084 [cs].}


\bibitem[\protect\citeauthoryear{Nyberg, Johnson, and Mehes}{Nyberg
  et~al\mbox{.}}{2022}]%
        {nyberg_cyber_2022}
\bibfield{author}{\bibinfo{person}{Jakob Nyberg}, \bibinfo{person}{Pontus
  Johnson}, {and} \bibinfo{person}{Andras Mehes}.}
  \bibinfo{year}{2022}\natexlab{}.
\newblock \showarticletitle{Cyber threat response using reinforcement learning
  in graph-based attack simulations}. In \bibinfo{booktitle}{{\em {NOMS}
  2022-2022 {IEEE}/{IFIP} {Network} {Operations} and {Management}
  {Symposium}}}. \bibinfo{publisher}{IEEE}, \bibinfo{address}{Budapest,
  Hungary}, \bibinfo{pages}{1--4}.
\newblock
\showISBNx{978-1-66540-601-7}
\showDOI{%
\url{https://doi.org/10.1109/NOMS54207.2022.9789835}}


\bibitem[\protect\citeauthoryear{{OpenAI}, Akkaya, Andrychowicz, Chociej,
  Litwin, McGrew, Petron, Paino, Plappert, Powell, Ribas, Schneider, Tezak,
  Tworek, Welinder, Weng, Yuan, Zaremba, and Zhang}{{OpenAI}
  et~al\mbox{.}}{2019}]%
        {openai_solving_2019}
\bibfield{author}{\bibinfo{person}{{OpenAI}}, \bibinfo{person}{Ilge Akkaya},
  \bibinfo{person}{Marcin Andrychowicz}, \bibinfo{person}{Maciek Chociej},
  \bibinfo{person}{Mateusz Litwin}, \bibinfo{person}{Bob McGrew},
  \bibinfo{person}{Arthur Petron}, \bibinfo{person}{Alex Paino},
  \bibinfo{person}{Matthias Plappert}, \bibinfo{person}{Glenn Powell},
  \bibinfo{person}{Raphael Ribas}, \bibinfo{person}{Jonas Schneider},
  \bibinfo{person}{Nikolas Tezak}, \bibinfo{person}{Jerry Tworek},
  \bibinfo{person}{Peter Welinder}, \bibinfo{person}{Lilian Weng},
  \bibinfo{person}{Qiming Yuan}, \bibinfo{person}{Wojciech Zaremba}, {and}
  \bibinfo{person}{Lei Zhang}.} \bibinfo{year}{2019}\natexlab{}.
\newblock \bibinfo{booktitle}{{\em Solving {Rubik}'s {Cube} with a {Robot}
  {Hand}}}.
\newblock


\bibitem[\protect\citeauthoryear{Perez-Nieves, Yang, Slumbers, Mguni, Wen, and
  Wang}{Perez-Nieves et~al\mbox{.}}{2021}]%
        {perez-nieves_modelling_2021}
\bibfield{author}{\bibinfo{person}{Nicolas Perez-Nieves},
  \bibinfo{person}{Yaodong Yang}, \bibinfo{person}{Oliver Slumbers},
  \bibinfo{person}{David~H Mguni}, \bibinfo{person}{Ying Wen}, {and}
  \bibinfo{person}{Jun Wang}.} \bibinfo{year}{2021}\natexlab{}.
\newblock \showarticletitle{Modelling behavioural diversity for learning in
  open-ended games}. In \bibinfo{booktitle}{{\em International conference on
  machine learning}}. \bibinfo{publisher}{PMLR}, \bibinfo{pages}{8514--8524}.
\newblock


\bibitem[\protect\citeauthoryear{Santucci, Oudeyer, Barto, and
  Baldassarre}{Santucci et~al\mbox{.}}{2020}]%
        {santucci_intrinsically_2020}
\bibfield{author}{\bibinfo{person}{Vieri~Giuliano Santucci},
  \bibinfo{person}{Pierre-Yves Oudeyer}, \bibinfo{person}{Andrew Barto}, {and}
  \bibinfo{person}{Gianluca Baldassarre}.} \bibinfo{year}{2020}\natexlab{}.
\newblock \bibinfo{title}{Intrinsically motivated open-ended learning in
  autonomous robots}.
\newblock   (\bibinfo{year}{2020}).
\newblock


\bibitem[\protect\citeauthoryear{Sharma, Gu, Levine, Kumar, and Hausman}{Sharma
  et~al\mbox{.}}{2019}]%
        {sharma_dynamics-aware_2019}
\bibfield{author}{\bibinfo{person}{Archit Sharma}, \bibinfo{person}{Shixiang
  Gu}, \bibinfo{person}{Sergey Levine}, \bibinfo{person}{Vikash Kumar}, {and}
  \bibinfo{person}{Karol Hausman}.} \bibinfo{year}{2019}\natexlab{}.
\newblock \bibinfo{title}{Dynamics-{Aware} {Unsupervised} {Discovery} of
  {Skills}}.
\newblock   (\bibinfo{year}{2019}).
\newblock
\showDOI{%
\url{https://doi.org/10.48550/ARXIV.1907.01657}}


\bibitem[\protect\citeauthoryear{Sharma, Gupta, Levine, Hausman, and
  Finn}{Sharma et~al\mbox{.}}{2021}]%
        {sharma_autonomous_2021}
\bibfield{author}{\bibinfo{person}{Archit Sharma}, \bibinfo{person}{Abhishek
  Gupta}, \bibinfo{person}{Sergey Levine}, \bibinfo{person}{Karol Hausman},
  {and} \bibinfo{person}{Chelsea Finn}.} \bibinfo{year}{2021}\natexlab{}.
\newblock \bibinfo{title}{Autonomous {Reinforcement} {Learning} via {Subgoal}
  {Curricula}}.
\newblock   (\bibinfo{year}{2021}).
\newblock
\showDOI{%
\url{https://doi.org/10.48550/ARXIV.2107.12931}}


\bibitem[\protect\citeauthoryear{Standen, Lucas, Bowman, Richer, Kim, and
  Marriott}{Standen et~al\mbox{.}}{2021}]%
        {standen_cyborg_2021}
\bibfield{author}{\bibinfo{person}{Maxwell Standen}, \bibinfo{person}{Martin
  Lucas}, \bibinfo{person}{David Bowman}, \bibinfo{person}{Toby~J Richer},
  \bibinfo{person}{Junae Kim}, {and} \bibinfo{person}{Damian Marriott}.}
  \bibinfo{year}{2021}\natexlab{}.
\newblock \showarticletitle{Cyborg: {A} gym for the development of autonomous
  cyber agents}.
\newblock \bibinfo{journal}{{\em arXiv preprint arXiv:2108.09118\/}}
  (\bibinfo{year}{2021}).
\newblock


\bibitem[\protect\citeauthoryear{Strom, Applebaum, Miller, Nickels, Pennington,
  and Thomas}{Strom et~al\mbox{.}}{2018}]%
        {strom_mitre_2018}
\bibfield{author}{\bibinfo{person}{Blake~E. Strom}, \bibinfo{person}{Andy
  Applebaum}, \bibinfo{person}{Douglas~P. Miller}, \bibinfo{person}{Kathryn~C.
  Nickels}, \bibinfo{person}{Adam~G. Pennington}, {and}
  \bibinfo{person}{Cody~B. Thomas}.} \bibinfo{year}{2018}\natexlab{}.
\newblock \showarticletitle{{MITRE} {ATT}\&{CK}™ : {Design} and
  {Philosophy}}.
\newblock  (\bibinfo{date}{July} \bibinfo{year}{2018}).
\newblock
\showURL{%
\url{https://www.mitre.org/publications/technical-papers/mitre-attack-design-and-philosophy}}


\bibitem[\protect\citeauthoryear{Tan}{Tan}{2022}]%
        {zhi_xuan_2022}
\bibfield{author}{\bibinfo{person}{Zhi-Xuan Tan}.}
  \bibinfo{year}{2022}\natexlab{}.
\newblock \bibinfo{title}{PDDL.jl: An Extensible Interpreter and Compiler
  Interface for Fast and Flexible AI Planning.}
\newblock   (\bibinfo{year}{2022}).
\newblock


\bibitem[\protect\citeauthoryear{Team, Seifert, Betser, Blum, Bono, and
  Farris}{Team et~al\mbox{.}}{2021a}]%
        {team_cyberbattlesim_2021}
\bibfield{author}{\bibinfo{person}{Microsoft Defender~Research Team},
  \bibinfo{person}{Christian Seifert}, \bibinfo{person}{Michael Betser},
  \bibinfo{person}{William Blum}, \bibinfo{person}{James Bono}, {and}
  \bibinfo{person}{Kate Farris}.} \bibinfo{year}{2021}\natexlab{a}.
\newblock \bibinfo{title}{{CyberBattleSim}}.
\newblock   (\bibinfo{year}{2021}).
\newblock
\showURL{%
\url{https://github.com/microsoft/cyberbattlesim}}


\bibitem[\protect\citeauthoryear{Team, Stooke, Mahajan, Barros, Deck, Bauer,
  Sygnowski, Trebacz, Jaderberg, Mathieu, McAleese, Bradley-Schmieg, Wong,
  Porcel, Raileanu, Hughes-Fitt, Dalibard, and Czarnecki}{Team
  et~al\mbox{.}}{2021b}]%
        {team_open-ended_2021}
\bibfield{author}{\bibinfo{person}{Open Ended~Learning Team},
  \bibinfo{person}{Adam Stooke}, \bibinfo{person}{Anuj Mahajan},
  \bibinfo{person}{Catarina Barros}, \bibinfo{person}{Charlie Deck},
  \bibinfo{person}{Jakob Bauer}, \bibinfo{person}{Jakub Sygnowski},
  \bibinfo{person}{Maja Trebacz}, \bibinfo{person}{Max Jaderberg},
  \bibinfo{person}{Michaël Mathieu}, \bibinfo{person}{Nat McAleese},
  \bibinfo{person}{Nathalie Bradley-Schmieg}, \bibinfo{person}{Nathaniel Wong},
  \bibinfo{person}{Nicolas Porcel}, \bibinfo{person}{Roberta Raileanu},
  \bibinfo{person}{Steph Hughes-Fitt}, \bibinfo{person}{Valentin Dalibard},
  {and} \bibinfo{person}{Wojciech~Marian Czarnecki}.}
  \bibinfo{year}{2021}\natexlab{b}.
\newblock \showarticletitle{Open-{Ended} {Learning} {Leads} to {Generally}
  {Capable} {Agents}}.
\newblock \bibinfo{journal}{{\em CoRR\/}}  \bibinfo{volume}{abs/2107.12808}
  (\bibinfo{year}{2021}).
\newblock
\showURL{%
\url{https://arxiv.org/abs/2107.12808}}


\bibitem[\protect\citeauthoryear{Tran, Akella, Standen, Kim, Bowman, Richer,
  and Lin}{Tran et~al\mbox{.}}{2021}]%
        {tran_deep_2021}
\bibfield{author}{\bibinfo{person}{Khuong Tran}, \bibinfo{person}{Ashlesha
  Akella}, \bibinfo{person}{Maxwell Standen}, \bibinfo{person}{Junae Kim},
  \bibinfo{person}{David Bowman}, \bibinfo{person}{Toby Richer}, {and}
  \bibinfo{person}{Chin-Teng Lin}.} \bibinfo{year}{2021}\natexlab{}.
\newblock \showarticletitle{Deep hierarchical reinforcement agents for
  automated penetration testing}.
\newblock \bibinfo{journal}{{\em arXiv preprint arXiv:2109.06449\/}}
  (\bibinfo{year}{2021}).
\newblock


\bibitem[\protect\citeauthoryear{Yang, Luo, Wen, Slumbers, Graves, Ammar, Wang,
  and Taylor}{Yang et~al\mbox{.}}{2021}]%
        {yang_diverse_2021}
\bibfield{author}{\bibinfo{person}{Yaodong Yang}, \bibinfo{person}{Jun Luo},
  \bibinfo{person}{Ying Wen}, \bibinfo{person}{Oliver Slumbers},
  \bibinfo{person}{Daniel Graves}, \bibinfo{person}{Haitham~Bou Ammar},
  \bibinfo{person}{Jun Wang}, {and} \bibinfo{person}{Matthew~E Taylor}.}
  \bibinfo{year}{2021}\natexlab{}.
\newblock \showarticletitle{Diverse auto-curriculum is critical for successful
  real-world multiagent learning systems}.
\newblock \bibinfo{journal}{{\em arXiv preprint arXiv:2102.07659\/}}
  (\bibinfo{year}{2021}).
\newblock


\bibitem[\protect\citeauthoryear{Zhou, Liu, Hou, Zhong, and Zhang}{Zhou
  et~al\mbox{.}}{2021}]%
        {zhou_autonomous_2021}
\bibfield{author}{\bibinfo{person}{Shicheng Zhou}, \bibinfo{person}{Jingju
  Liu}, \bibinfo{person}{Dongdong Hou}, \bibinfo{person}{Xiaofeng Zhong}, {and}
  \bibinfo{person}{Yue Zhang}.} \bibinfo{year}{2021}\natexlab{}.
\newblock \showarticletitle{Autonomous penetration testing based on improved
  deep q-network}.
\newblock \bibinfo{journal}{{\em Applied Sciences\/}} \bibinfo{volume}{11},
  \bibinfo{number}{19} (\bibinfo{year}{2021}), \bibinfo{pages}{8823}.
\newblock


\bibitem[\protect\citeauthoryear{Zhu, Chen, Zhu, and Zhu}{Zhu
  et~al\mbox{.}}{2024}]%
        {zhu_effective_2024}
\bibfield{author}{\bibinfo{person}{Zhengwei Zhu}, \bibinfo{person}{Miaojie
  Chen}, \bibinfo{person}{Chenyang Zhu}, {and} \bibinfo{person}{Yanping Zhu}.}
  \bibinfo{year}{2024}\natexlab{}.
\newblock \showarticletitle{Effective defense strategies in network security
  using improved double dueling deep {Q}-network}.
\newblock \bibinfo{journal}{{\em Computers \& Security\/}}
  \bibinfo{volume}{136} (\bibinfo{date}{Jan.} \bibinfo{year}{2024}),
  \bibinfo{pages}{103578}.
\newblock
\showISSN{01674048}
\showDOI{%
\url{https://doi.org/10.1016/j.cose.2023.103578}}


\end{thebibliography}

\section{Appendix}
\label{appendix1}

\subsection{Reward Functions}
In order to guide the behavior of a RL network defender to align with defense
and quality of service goals, it is crucial to carefully select a reward
function for the task at hand. Choosing an appropriate reward function is a
critical aspect of the RL design process and can greatly impact the success of
the network defender. For the results in Section \ref{evaluation} we applied a sparse reward that depends on the goal and metric specified in PDDL. However, we explored other options as well, including:
\begin{itemize}
    \item An outcome-based sparse reward that returns -1
    if the attacker agent compromises the network. 1 if the attacker does
    not compromise the network by the end of the episode, 0 otherwise. 
    \item The QoS-aware sparse reward is similar to the  
    outcome-based-sparse reward but also accounts for 
    Quality of Service (QoS) by penalizing at the end of the episode 
    based on the number of CJ relocation attempts, host isolation attempts, 
    as well as host reported bad QoS events in the network (e.g scp\_failed, ssh\_failed, http
    \_failed). This reward has range in the interval $(0, 8]$.
    \item  The scaled-QoS-aware sparse reward is a a min-max scaled version of the QoS-aware sparse reward with range in the interval $(0, 1]$. 
\end{itemize}

It has been shown that optimal behavior may be achieved via multiple rewards
\cite{ng_1999}. However, in practice, some of them result in substantial
reductions in learning time \cite{sowerby2022designing}. As we described in
Section \ref{subsection:task_representation}, our reward function aims to be
consistent across the space of possible tasks exposed to the RL agent.
Therefore, we designed reward functions that are \emph{invariant} to: network
size, red agent TTP, and amount network traffic. 

\subsubsection{Dense rewards}
We did not extensively explore dense reward functions in this work. In practice,
dense rewards that are nonzero on every state must be carefully designed to
perform well. Poorly designed dense reward functions have been shown to perform
worse than goal sparse reward functions \cite{sowerby2022designing}. More
importantly, as the number of tasks to master increases (i.e., as is the case in
Open-ended Learning), it is simpler to implement sparse reward functions. In
future work we explore the use of other types of less sparse rewards to
accommodate interactions with human operators and to formulate long-horizon
defense tasks as sequences of smaller sub tasks.

\subsection{PDDL Terms for Goals and Metrics}
\label{appendix:pddl_terms}

Figures \ref{fig:goal-metrics1-12} - \ref{fig:goal-metrics37-43} list the goal-metric pairs that we used in our evaluation. These use the following terms:

\begin{itemize}
\item red-inactive: A boolean that is true when the red is inactive
\item declared-victory: A boolean that is true when the red agent declares victory
\item real-compromise: A boolean that is true when the red agent compromises a real asses (as opposed to a fake one)
\item good-qos-events: An integer that corresponds to the number of good QoS events
\item steps-to-survive: An integer that corresponds to the upper bound of necessary steps for the red agent to succeed without intervention from the blue agent
\item bad-qos-events: An integer that corresponds to the number of bad QoS events
\item isolate-actions: An integer that corresponds to the number of actions to isolate hosts
\item crown-jewel-relocations: An integer that corresponds to the number of actions to relocate a crown jewel
\item num-blue-actions: An integer that corresponds to the total number of blue actions
\item worst-contributor-isolations: An integer that corresponds to the number of actions attempting to isolate a suspicious host
\item crown-jewel-isolations: An integer that corresponds to the number of actions attempting to isolate a crown jewel 
\item worst-contributor-reimages: An integer that corresponds to the number of actions attempting to reimage a suspicious host
\item worst-contributor-relocations: An integer that corresponds to the number of actions attempting to relocate a suspicious host to a different subnet
\item worst-contributor-honeys: An integer that corresponds to the number of actions attempting to relocate a suspicious host to a honey network
\item full-nontrivial-blue-actions: An integer that corresponds to the number of all blue actions that "do something"
\item nontrivial-blue-actions: An integer that corresponds to the number of represented blue actions that "do something"
\end{itemize}

\begin{figure*}[t!]
    \centering
    
\begin{lstlisting}[
    basicstyle=\tiny, %or \small or \footnotesize etc.
    breaklines=true
]
{
    "id": 1,
    "original_id": 1,
    "goal": "(:goal (> nontrivial-blue-actions 0))",
    "metric": "(:metric minimize (0))"
},
{
    "id": 2,
    "original_id": 2,
    "goal": "(:goal (> worst-contributor-isolations 0))",
    "metric": "(:metric minimize (0))"
},
{
    "id": 3,
    "original_id": 3,
    "goal": "(:goal (> worst-contributor-isolations 0))",
    "metric": "(:metric minimize (/ nontrivial-blue-actions steps-to-survive))"
},
{
    "id": 4,
    "original_id": 4,
    "goal": "(:goal (> worst-contributor-isolations 0))",
    "metric": "(:metric minimize (qos-penalty))"
},
{
    "id": 5,
    "original_id": 5,
    "goal": "(:goal (> worst-contributor-isolations 0))",
    "metric": "(:metric minimize (+ (/ nontrivial-blue-actions steps-to-survive) (qos-penalty)) )"
},
{
    "id": 6,
    "original_id": 6,
    "goal": "(:goal (> crown-jewel-isolations 0))",
    "metric": "(:metric minimize (0))"
},
{
    "id": 7,
    "original_id": 7,
    "goal": "(:goal (> crown-jewel-isolations 0))",
    "metric": "(:metric minimize (/ nontrivial-blue-actions steps-to-survive))"
},
{
    "id": 8,
    "original_id": 8,
    "goal": "(:goal (> crown-jewel-isolations 0))",
    "metric": "(:metric minimize (qos-penalty))"
},
{
    "id": 9,
    "original_id": 9,
    "goal": "(:goal (> crown-jewel-isolations 0))",
    "metric": "(:metric minimize (+ (/ nontrivial-blue-actions steps-to-survive) (qos-penalty)))"
},
{
    "id": 10,
    "original_id": 10,
    "goal": "(:goal (> worst-contributor-reimages 0))",
    "metric": "(:metric minimize (0))"
},
{
    "id": 11,
    "original_id": 11,
    "goal": "(:goal (> worst-contributor-reimages 0))",
    "metric": "(:metric minimize (/ nontrivial-blue-actions steps-to-survive))"
},
{
    "id": 12,
    "original_id": 12,
    "goal": "(:goal (> worst-contributor-reimages 0))",
    "metric": "(:metric minimize (qos-penalty))"
},

\end{lstlisting}
\caption{Goal-metrics 1-12.}
\label{fig:goal-metrics1-12}
\end{figure*}

\begin{figure*}[t!]
    \centering
    
\begin{lstlisting}[
    basicstyle=\tiny, %or \small or \footnotesize etc.
    breaklines=true
]

{
    "id": 13,
    "original_id": 13,
    "goal": "(:goal (> worst-contributor-reimages 0))",
    "metric": "(:metric minimize (+ (/ nontrivial-blue-actions steps-to-survive) (qos-penalty)))"
},
{
    "id": 14,
    "original_id": 14,
    "goal": "(:goal (> worst-contributor-relocations 0))",
    "metric": "(:metric minimize (0))"
},
{
    "id": 15,
    "original_id": 15,
    "goal": "(:goal (> worst-contributor-relocations 0))",
    "metric": "(:metric minimize (/ nontrivial-blue-actions steps-to-survive))"
},
{
    "id": 16,
    "original_id": 16,
    "goal": "(:goal (> worst-contributor-relocations 0))",
    "metric": "(:metric minimize (qos-penalty))"
},
{
    "id": 17,
    "original_id": 17,
    "goal": "(:goal (> worst-contributor-relocations 0))",
    "metric": "(:metric minimize (+ (/ nontrivial-blue-actions steps-to-survive) (qos-penalty)))"
},
{
    "id": 18,
    "original_id": 18,
    "goal": "(:goal (> worst-contributor-honeys 0))",
    "metric": "(:metric minimize (0))"
},
{
    "id": 19,
    "original_id": 19,
    "goal": "(:goal (> worst-contributor-honeys 0))",
    "metric": "(:metric minimize (/ nontrivial-blue-actions steps-to-survive))"
},
{
    "id": 20,
    "original_id": 20,
    "goal": "(:goal (> worst-contributor-honeys 0))",
    "metric": "(:metric minimize (qos-penalty))"
},
{
    "id": 21,
    "original_id": 21,
    "goal": "(:goal (> worst-contributor-honeys 0))",
    "metric": "(:metric minimize (+ (/ nontrivial-blue-actions steps-to-survive) (qos-penalty)))"
},
{
    "id": 22,
    "original_id": 22,
    "goal": "(:goal (and (not real-compromise) (> worst-contributor-isolations 0)))",
    "metric": "(:metric minimize (0))"
},
{
    "id": 23,
    "original_id": 23,
    "goal": "(:goal (and (not real-compromise) (> worst-contributor-isolations 0)))",
    "metric": "(:metric minimize (/ nontrivial-blue-actions steps-to-survive))"
},
{
    "id": 24,
    "original_id": 24,
    "goal": "(:goal (and (not real-compromise) (> worst-contributor-isolations 0)))",
    "metric": "(:metric minimize (qos-penalty))"
},
\end{lstlisting}
\caption{Goal-metrics 13-24.}
\label{fig:goal-metrics13-24}
\end{figure*}

\begin{figure*}[t!]
    \centering
    
\begin{lstlisting}[
    basicstyle=\tiny, %or \small or \footnotesize etc.
    breaklines=true
]
{
    "id": 25,
    "original_id": 25,
    "goal": "(:goal (and (not real-compromise) (> worst-contributor-isolations 0)))",
    "metric": "(:metric minimize (+ (/ nontrivial-blue-actions steps-to-survive) (qos-penalty)) )"
},
{
    "id": 26,
    "original_id": 26,
    "goal": "(:goal (and (not real-compromise) (> crown-jewel-isolations 0)))",
    "metric": "(:metric minimize (0))"
},
{
    "id": 27,
    "original_id": 27,
    "goal": "(:goal (and (not real-compromise) (> crown-jewel-isolations 0)))",
    "metric": "(:metric minimize (/ nontrivial-blue-actions steps-to-survive))"
},
{
    "id": 28,
    "original_id": 28,
    "goal": "(:goal (and (not real-compromise) (> crown-jewel-isolations 0)))",
    "metric": "(:metric minimize (qos-penalty))"
},
{
    "id": 29,
    "original_id": 29,
    "goal": "(:goal (and (not real-compromise) (> crown-jewel-isolations 0)))",
    "metric": "(:metric minimize (+ (/ nontrivial-blue-actions steps-to-survive) (qos-penalty)))"
},
{
    "id": 30,
    "original_id": 30,
    "goal": "(:goal (and (not real-compromise) (> worst-contributor-reimages 0)))",
    "metric": "(:metric minimize (0))"
},
{
    "id": 31,
    "original_id": 31,
    "goal": "(:goal (and (not real-compromise) (> worst-contributor-reimages 0)))",
    "metric": "(:metric minimize (/ nontrivial-blue-actions steps-to-survive))"
},
{
    "id": 32,
    "original_id": 32,
    "goal": "(:goal (and (not real-compromise) (> worst-contributor-reimages 0)))",
    "metric": "(:metric minimize (qos-penalty))"
},
{
    "id": 33,
    "original_id": 33,
    "goal": "(:goal (and (not real-compromise) (> worst-contributor-reimages 0)))",
    "metric": "(:metric minimize (+ (/ nontrivial-blue-actions steps-to-survive) (qos-penalty)))"
},
{
    "id": 34,
    "original_id": 34,
    "goal": "(:goal (and (not real-compromise) (> worst-contributor-relocations 0)))",
    "metric": "(:metric minimize (0))"
},
{
    "id": 35,
    "original_id": 35,
    "goal": "(:goal (and (not real-compromise) (> worst-contributor-relocations 0)))",
    "metric": "(:metric minimize (/ nontrivial-blue-actions steps-to-survive))"
},
{
    "id": 36,
    "original_id": 36,
    "goal": "(:goal (and (not real-compromise) (> worst-contributor-relocations 0)))",
    "metric": "(:metric minimize (qos-penalty))"
},
\end{lstlisting}
\caption{Goal-metrics 25-36.}
\label{fig:goal-metrics25-36}
\end{figure*}

\begin{figure*}[t!]
    \centering
    
\begin{lstlisting}[
    basicstyle=\tiny, %or \small or \footnotesize etc.
    breaklines=true
]
{
    "id": 37,
    "original_id": 37,
    "goal": "(:goal (and (not real-compromise) (> worst-contributor-relocations 0)))",
    "metric": "(:metric minimize (+ (/ nontrivial-blue-actions steps-to-survive) (qos-penalty)))"
},
{
    "id": 38,
    "original_id": 38,
    "goal": "(:goal (and (not real-compromise) (> worst-contributor-honeys 0)))",
    "metric": "(:metric minimize (0))"
},
{
    "id": 39,
    "original_id": 39,
    "goal": "(:goal (and (not real-compromise) (> worst-contributor-honeys 0)))",
    "metric": "(:metric minimize (/ nontrivial-blue-actions steps-to-survive))"
},
{
    "id": 40,
    "original_id": 40,
    "goal": "(:goal (and (not real-compromise) (> worst-contributor-honeys 0)))",
    "metric": "(:metric minimize (qos-penalty))"
},
{
    "id": 41,
    "original_id": 41,
    "goal": "(:goal (and (not real-compromise) (> worst-contributor-honeys 0)))",
    "metric": "(:metric minimize (+ (/ nontrivial-blue-actions steps-to-survive) (qos-penalty)))"
},
{
    "id": 42,
    "original_id": 42,
    "goal": "(:goal (and (not real-compromise) (red-declared-victory)))",
    "metric": "(:metric minimize (0))"
},
{
    "id": 43,
    "original_id": 43,
    "goal": "(:goal (and (not real-compromise) (red-declared-victory)))",
    "metric": "(:metric minimize (+ (/ nontrivial-blue-actions steps-to-survive) (qos-penalty)))"
}
\end{lstlisting}
\caption{Goal-metrics 37-43.}
\label{fig:goal-metrics37-43}
\end{figure*}

\begin{figure*}[t!]
    \centering
    \subfigure{\includegraphics[width=.55\textwidth, height=.25\textheight]{plots/scaling/reward.png}}
\begin{tabular}{ccc}
    \includegraphics[trim={0 0 3.1cm 0},clip,width=.3\textwidth]{plots/scaling/qos_penalty.png} &
    \includegraphics[trim={0 0 3.1cm 0},clip,width=.3\textwidth]{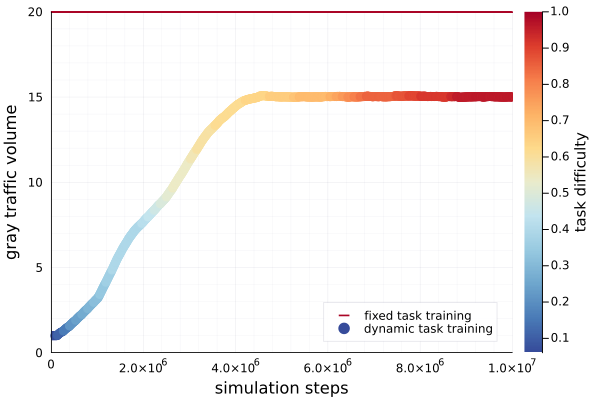} &
    \includegraphics[trim={0 0 3.1cm 0},clip,width=.3\textwidth]{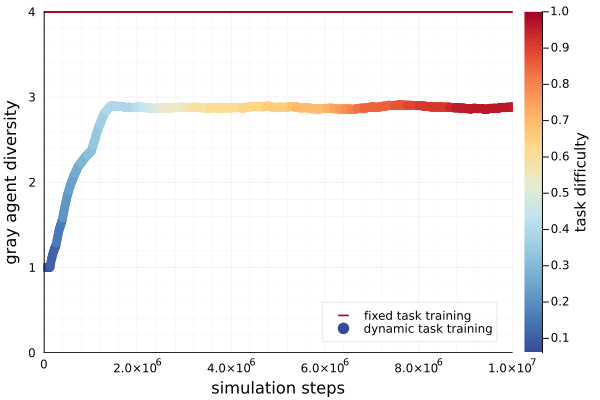} \\
    \includegraphics[trim={0 0 3.1cm 0},clip,width=.3\textwidth]{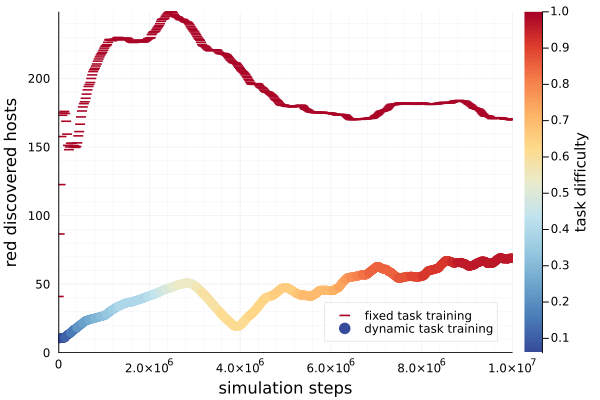} &
    \includegraphics[trim={0 0 3.1cm 0},clip,width=.3\textwidth]{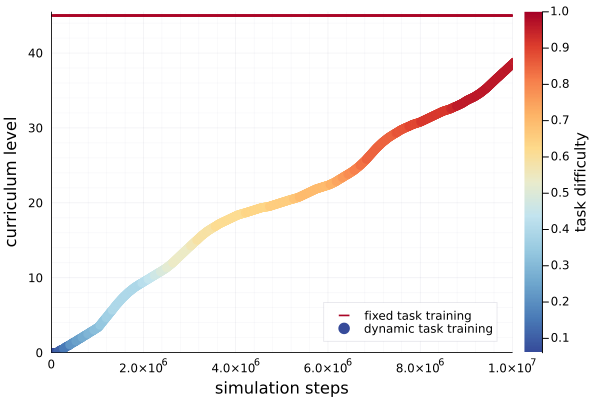} &
    \includegraphics[trim={0 0 3.1cm 0},clip,width=.3\textwidth]{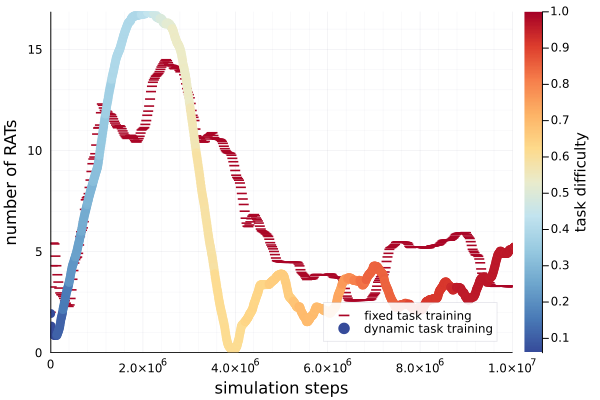} \\
    \includegraphics[trim={0 0 3.1cm 0},clip,width=.3\textwidth]{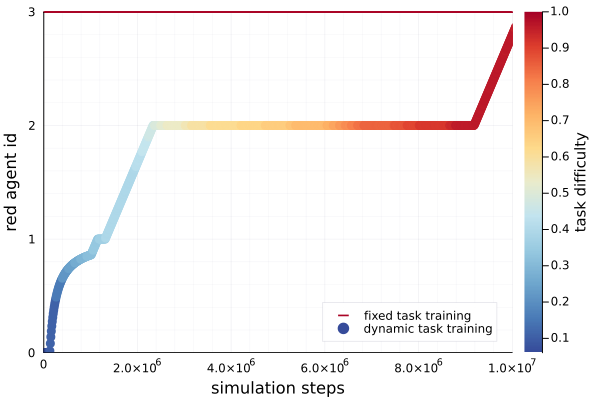} &
    \includegraphics[trim={0 0 3.1cm 0},clip,width=.3\textwidth]{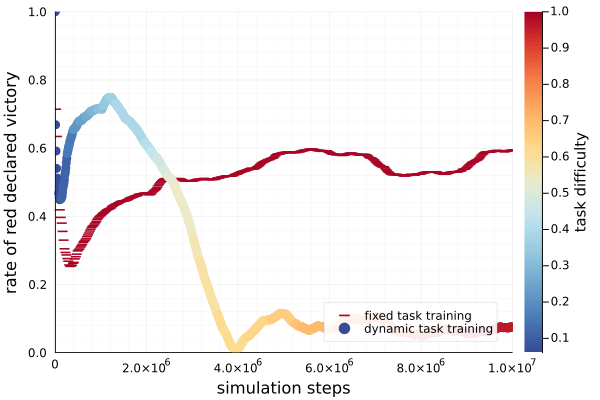} &
    \includegraphics[trim={0 0 3.1cm 0},clip,width=.3\textwidth]{plots/scaling/red_compromised_net.png} 
\end{tabular}
\caption{Performance comparison of three policies. One policy corresponds to
training with the fixed task selection strategy in Section~\ref{subsec:scaling}.
Another policy corresponds to training with dynamic task selection, also
described in Section \ref{subsec:scaling}. For reference, we compare the
performance of these policies with a \emph{trivial policy} in which the blue
agent does nothing. Note that it is possible to achieve worse performance than the
trivial policy, if in addition to allowing security compromises, the QoS was
degraded. The color of the markers represents the
difficulty of the task during the corresponding training iteration, derived from
the number of hosts, subnets, and variability of red and gray behaviors.}
\label{fig:fixed_vs_dynamic_full}
\end{figure*}

\begin{figure*}[t!]
    \centering
    \subfigure{\includegraphics[width=.55\textwidth, height=.25\textheight]{plots/task_selection/reward.png}}
\begin{tabular}{ccc}
    \includegraphics[trim={0 0 3cm 0},clip,width=.3\textwidth]{plots/task_selection/qos_penalty.png} &
    \includegraphics[trim={0 0 3cm 0},clip,width=.3\textwidth]{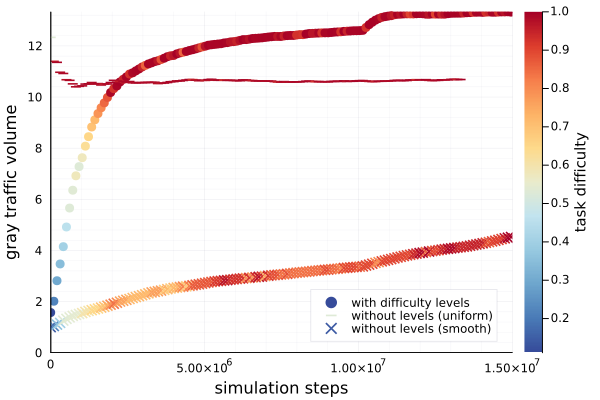} &
    \includegraphics[trim={0 0 3cm 0},clip,width=.3\textwidth]{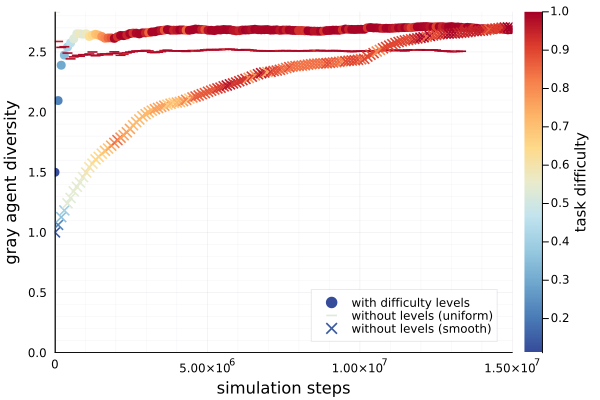} \\
    \includegraphics[trim={0 0 3cm 0},clip,width=.3\textwidth]{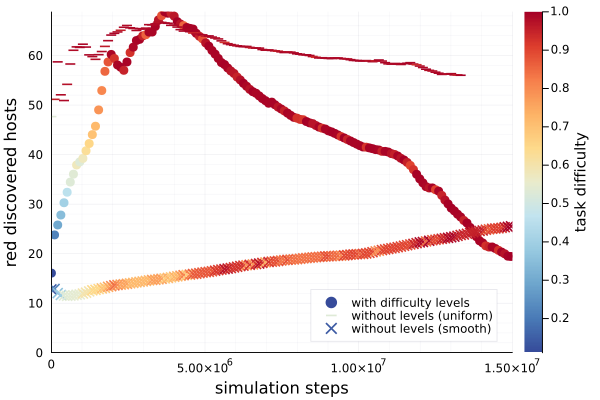} &
    \includegraphics[trim={0 0 3cm 0},clip,width=.3\textwidth]{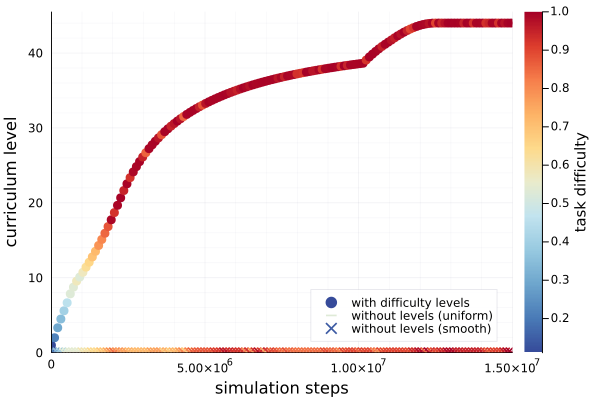} &
    \includegraphics[trim={0 0 3cm 0},clip,width=.3\textwidth]{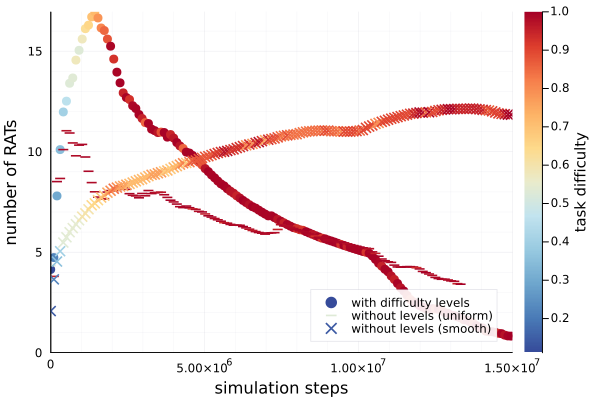} \\
    \includegraphics[trim={0 0 3cm 0},clip,width=.3\textwidth]{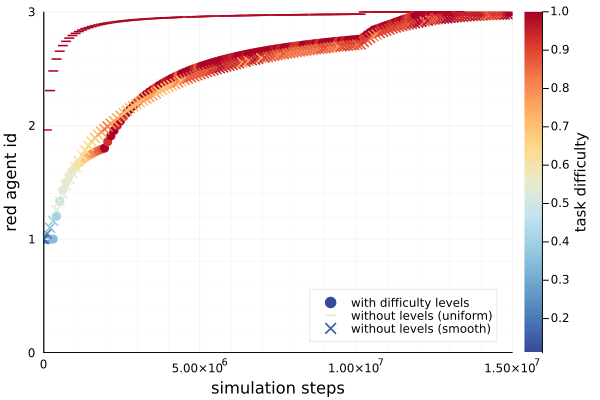} &
    \includegraphics[trim={0 0 3cm 0},clip,width=.3\textwidth]{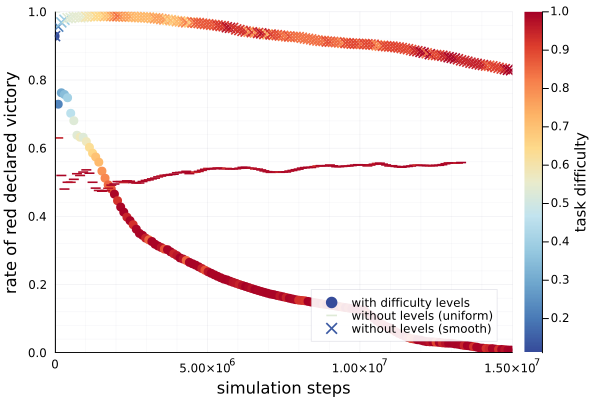} &
    \includegraphics[trim={0 0 3cm 0},clip,width=.3\textwidth]{plots/task_selection/red_compromised_net.png} 
\end{tabular}
\caption{Performance comparison of multiple task selection strategies. One
strategy uses difficulty levels to drive selection of tasks with increasing
difficulty. Other task selection strategies include choosing tasks uniformly at
random and choosing tasks via small variations (smoothly). While the average
reward for ``smooth'' task updates is higher, the difficulty of the task mastered
is higher with the approach that uses difficulty levels. The color of the markers
represents the difficulty of the task during the corresponding training
iteration, derived from the number of hosts, subnets, and variability of red and
gray behaviors.}
\label{fig:figsamplingstrategies_full}
\end{figure*}

\begin{figure*}[t!]
    \centering
    \subfigure{\includegraphics[width=.55\textwidth, height=.25\textheight]{plots/robustness/reward.png}}
\begin{tabular}{ccc}
    \includegraphics[trim={0 0 3.1cm 0},clip,width=.3\textwidth]{plots/robustness/qos_summary.png} &
    \includegraphics[trim={0 0 3.1cm 0},clip,width=.3\textwidth]{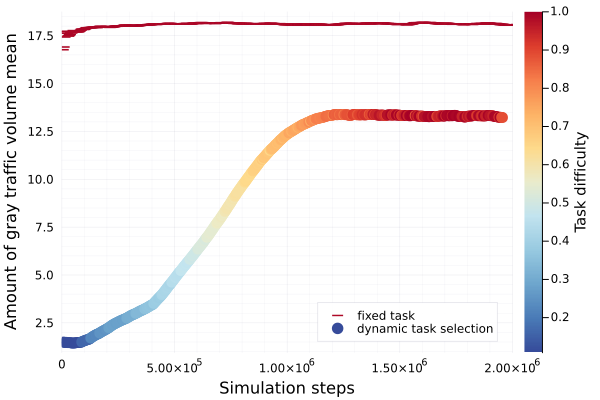} &
    \includegraphics[trim={0 0 3.1cm 0},clip,width=.3\textwidth]{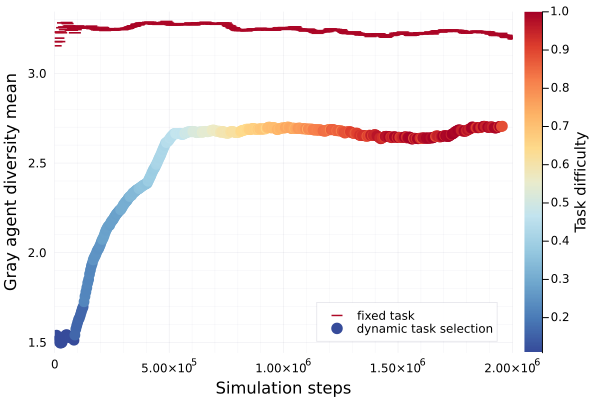} \\
    \includegraphics[trim={0 0 3.1cm 0},clip,width=.3\textwidth]{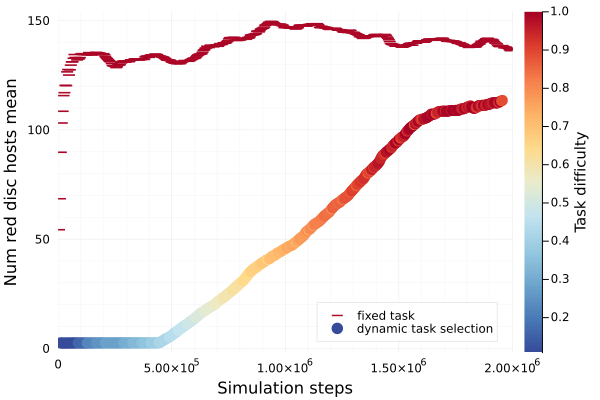} &
    \includegraphics[trim={0 0 3.1cm 0},clip,width=.3\textwidth]{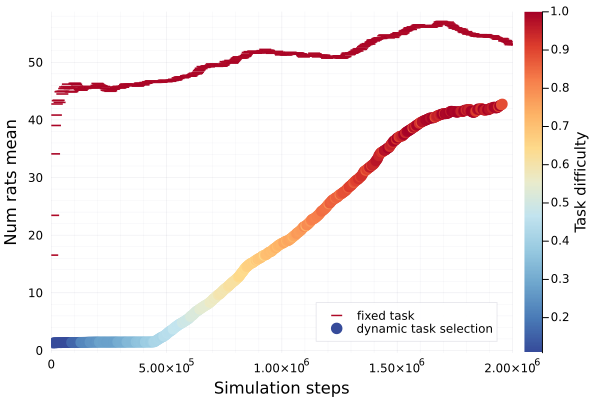} &
    \includegraphics[trim={0 0 3.1cm 0},clip,width=.3\textwidth]{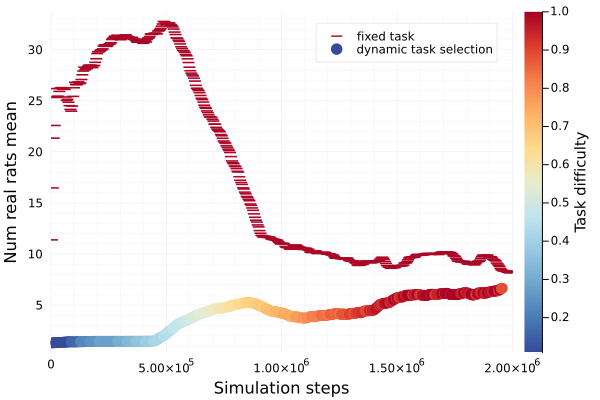} \\
    \includegraphics[trim={0 0 3.1cm 0},clip,width=.3\textwidth]{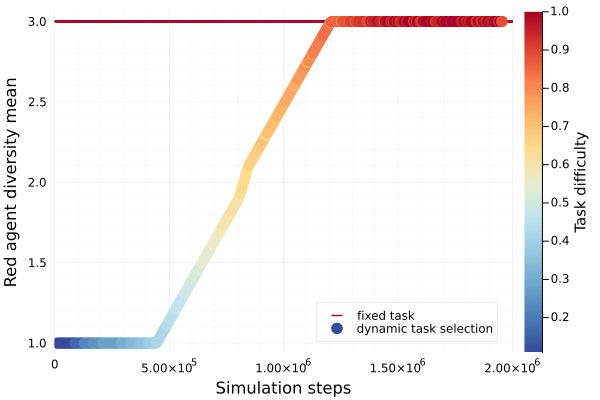} &
    \includegraphics[trim={0 0 3.1cm 0},clip,width=.3\textwidth]{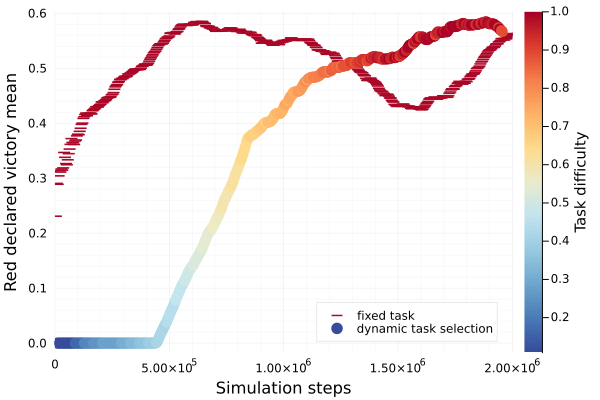} &
    \includegraphics[trim={0 0 3.1cm 0},clip,width=.3\textwidth]{plots/robustness/red_compromised.png} 
\end{tabular}
\caption{Performance comparison of two policies. One policy was trained via a
fixed task, with the goal of mitigating a traditional exfiltration attack. The
other one was trained via dynamic task selection, where the blue agent learns to
mitigate an exfiltration attack, where the red agent interleaves gray-like
actions between traditional red actions, with some probability. The color of the markers
represents the difficulty of the task during the corresponding training
iteration, derived from the number of hosts, subnets, and variability of red and
gray behaviors.}
\label{fig:fig_morphing_full}
\end{figure*}

\begin{figure*}[t!]
    \centering
    \subfigure{\includegraphics[width=.55\textwidth, height=.25\textheight]{plots/action_representations/reward.png}}
\begin{tabular}{ccc}
    \includegraphics[trim={0 0 3.1cm 0},clip,width=.3\textwidth]{plots/action_representations/qos_penalty.png} &
    \includegraphics[trim={0 0 3.1cm 0},clip,width=.3\textwidth]{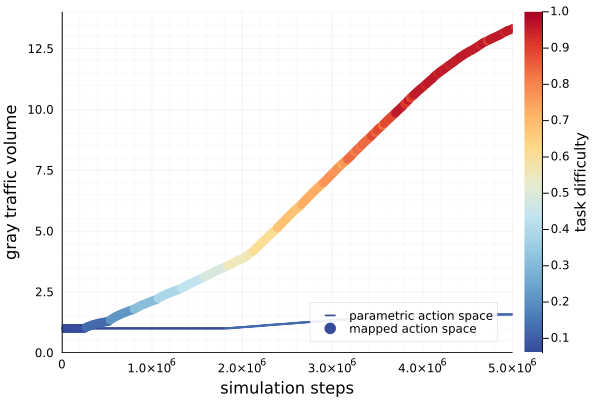} &
    \includegraphics[trim={0 0 3.1cm 0},clip,width=.3\textwidth]{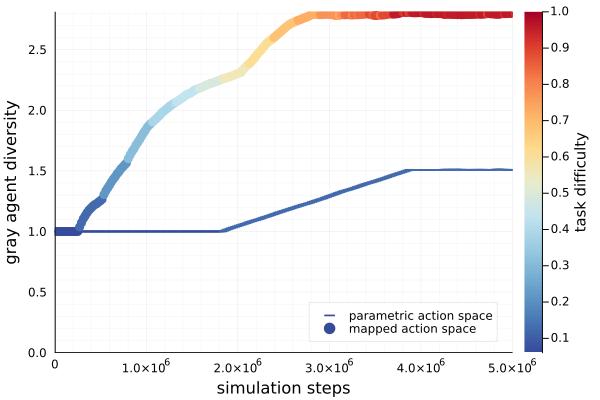} \\
    \includegraphics[trim={0 0 3.1cm 0},clip,width=.3\textwidth]{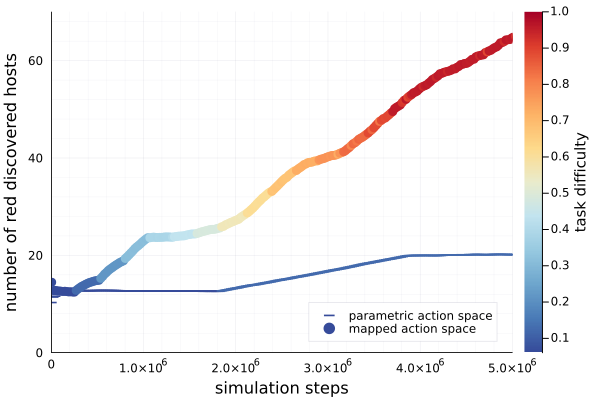} &
    \includegraphics[trim={0 0 3.1cm 0},clip,width=.3\textwidth]{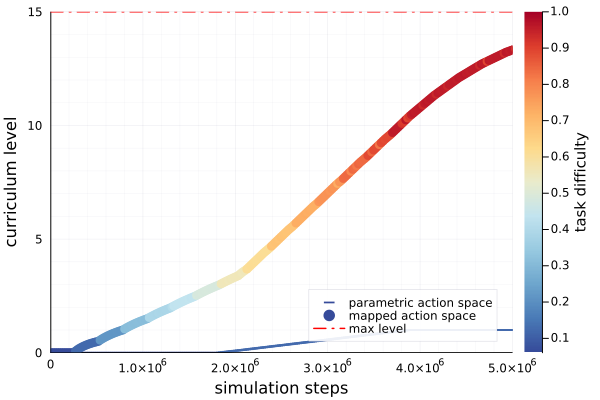} &
    \includegraphics[trim={0 0 3.1cm 0},clip,width=.3\textwidth]{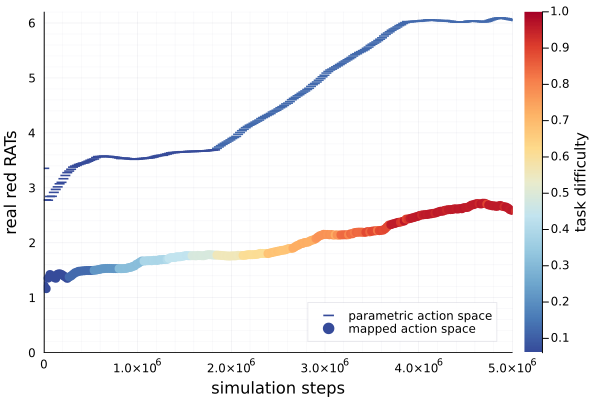} \\
    \includegraphics[trim={0 0 3.1cm 0},clip,width=.3\textwidth]{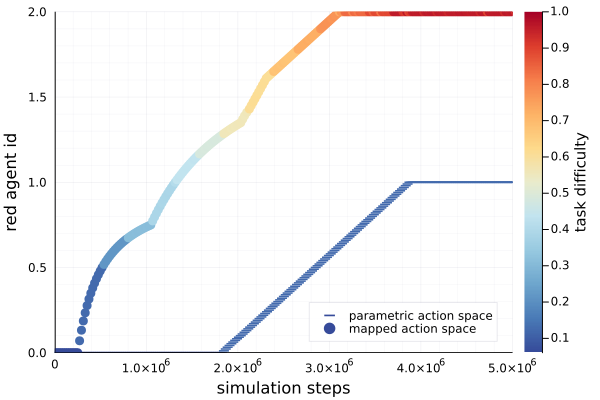} &
    \includegraphics[trim={0 0 3.1cm 0},clip,width=.3\textwidth]{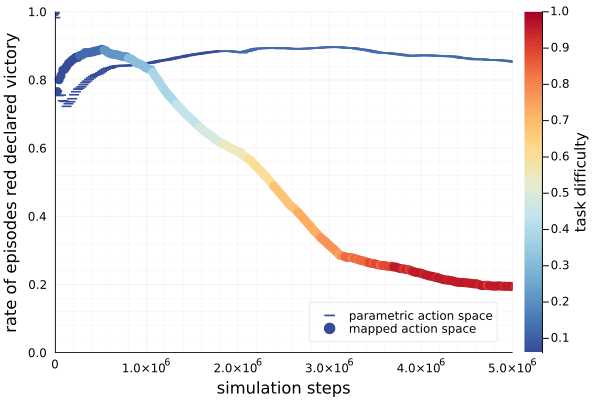} &
    \includegraphics[trim={0 0 3.1cm 0},clip,width=.3\textwidth]{plots/action_representations/red_compromised.png} 
\end{tabular}
\caption{Comparison in learning performance between two learning strategies that
only differ in the actions available to the agent. One learning strategy uses
action representations while the other allows the agent to select actions that
include the target host to execute the action over. Learning stagnates when the
action space is too large. In both cases, observation spaces, reward functions,
and network dynamics are the same.}
\label{fig:action_representations_full}
\end{figure*}

\begin{figure*}[t!]
    \centering
    \subfigure{\includegraphics[width=.55\textwidth, height=.25\textheight]{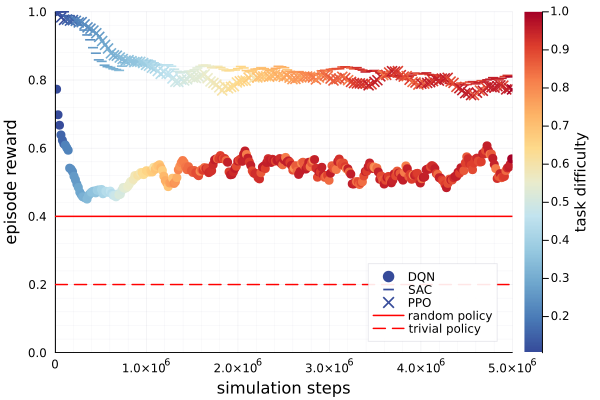}}
\begin{tabular}{ccc}
    \includegraphics[trim={0 0 3.1cm 0},clip,width=.3\textwidth]{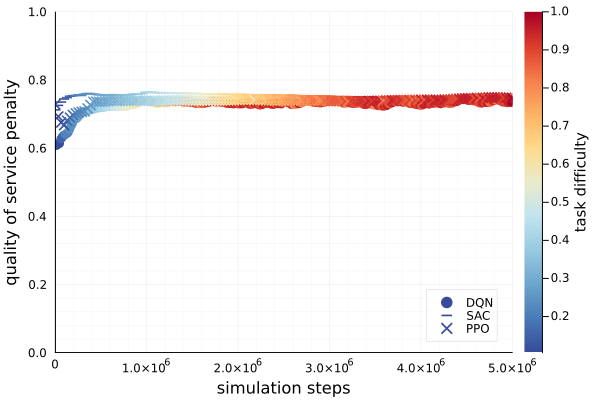} &
    \includegraphics[trim={0 0 3.1cm 0},clip,width=.3\textwidth]{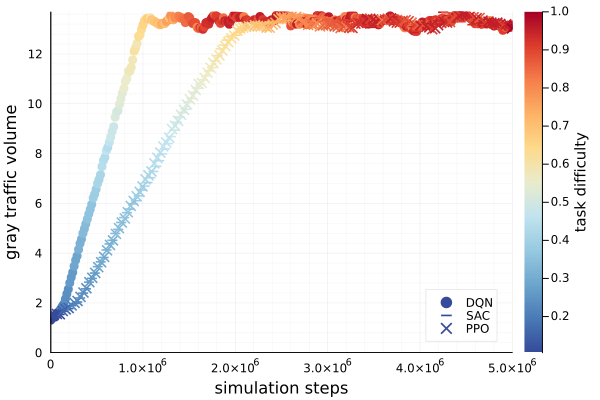} &
    \includegraphics[trim={0 0 3.1cm 0},clip,width=.3\textwidth]{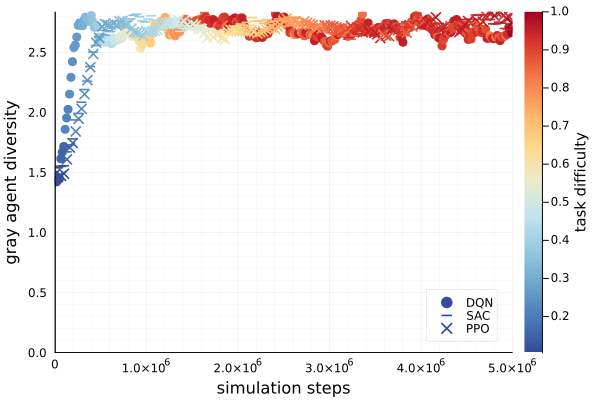} \\
    \includegraphics[trim={0 0 3.1cm 0},clip,width=.3\textwidth]{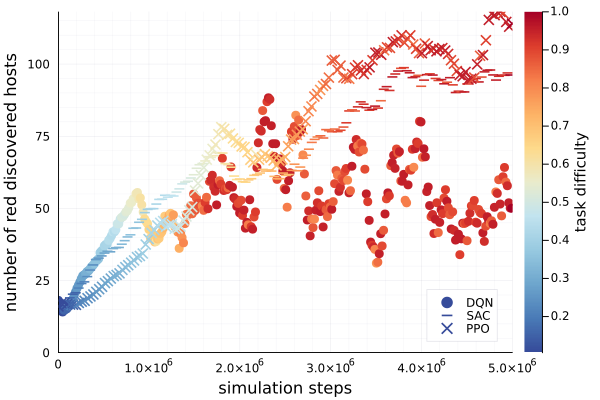} &
    \includegraphics[trim={0 0 3.1cm 0},clip,width=.3\textwidth]{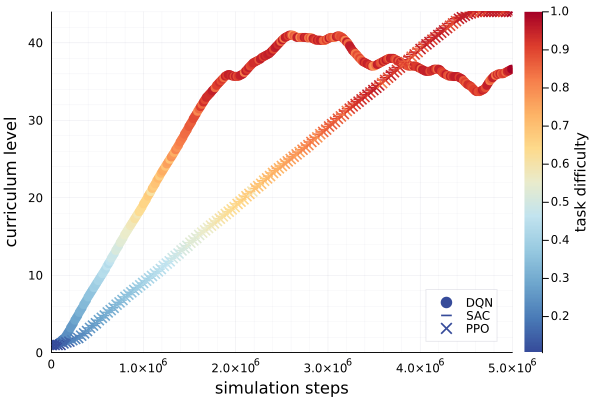} &
    \includegraphics[trim={0 0 3.1cm 0},clip,width=.3\textwidth]{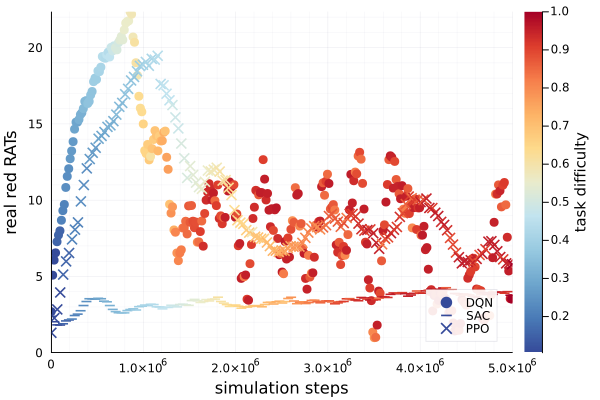} \\
    \includegraphics[trim={0 0 3.1cm 0},clip,width=.3\textwidth]{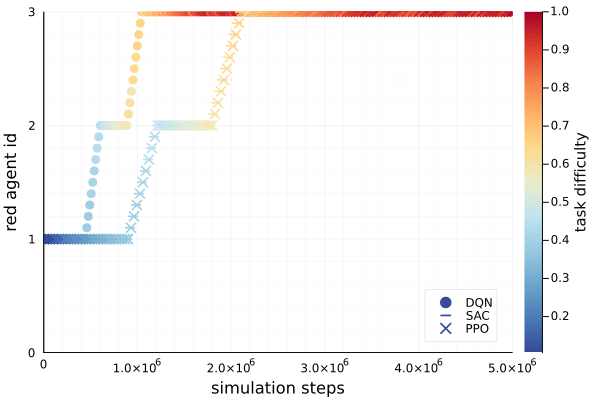} &
    \includegraphics[trim={0 0 3.1cm 0},clip,width=.3\textwidth]{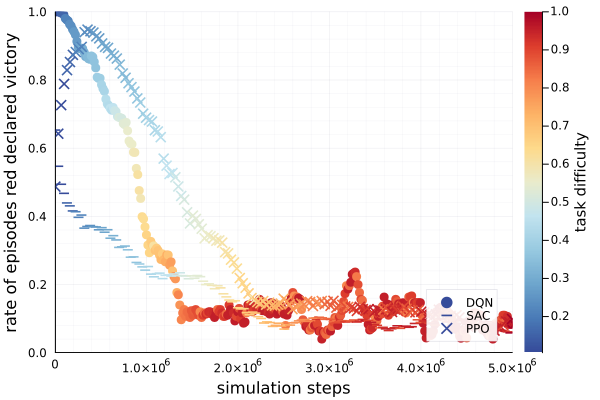} &
    \includegraphics[trim={0 0 3.1cm 0},clip,width=.3\textwidth]{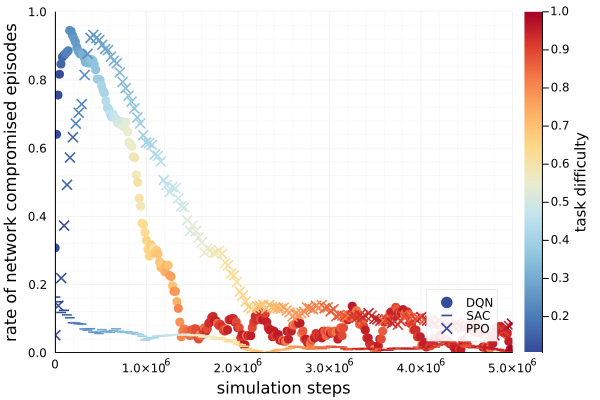} 
\end{tabular}
\caption{Performance comparison of algorithms with the same dynamic task selection strategy.}
\label{fig:fig_algorithms}
\end{figure*}

\subsection{Curriculum Design}

\preliminarytext{Though the challenge of architecting a good curriculum is a research area in its own right, we have a few practical findings from our experience designing a curriculum.}

\paragraph{Tier-Based Structuring of Goals.}
Training independent policies for each goal proved to be an efficient way to establish tiers of difficulty. Our 43 goals naturally divided into four phases:
\begin{itemize}
    \item \textbf{Initialization:} Goal 1
    \item \textbf{Basic Skills:} Goals 2--21
    \item \textbf{Applied Skills:} Goals 22--41
    \item \textbf{Advanced Goals:} Goals 42--43
\end{itemize}
\preliminarytext{In this context, skills refer to goals that focus the agent on using specific actions either for their own sake (\textit{basic skills}) or to achieve a major theme of the curriculum (\textit{applied skills}). Using this goal order resulted in an average reward of 0.75 and a compromise rate of 10\%, significantly outperforming the reverse goal order, which failed to progress through the curriculum.}

\paragraph{Importance of Ascending Difficulty.}
\preliminarytext{Structuring curriculums with ascending difficulty is essential, as front-loading too many challenges can overwhelm agents. For instance:}
\begin{itemize}
    \item \preliminarytext{A curriculum with multiple versions of a goal arranged by ascending difficulty outperforms using just the goal with the most difficult metric, improving average reward by 26\%.}
    \item \preliminarytext{Curriculums that include both basic and applied skills achieved a 15\% higher average reward than those without skill differentiation.}
\end{itemize}

\paragraph{Handling Atypical Adversaries.}
\preliminarytext{Starting curriculums with exclusive exposure to atypical adversaries (e.g., the inactive red agent) can completely derail training. Agents exposed to adversaries in this way developed policies that failed to generalize to the main set of adversaries, leading to cycles of demotion and promotion when encountering their first typical adversary. To mitigate this, it is best to heterogeneously mix experiences with atypical adversaries throughout training. This approach improved average reward by 10--15\%.}

\vspace{12cm}









\end{document}